\documentclass[journal]{IEEEtran}
\usepackage{amsmath,amsfonts}
\usepackage{algorithmic}
\usepackage{algorithm}
\usepackage{array}
\usepackage[caption=false,font=normalsize,labelfont=sf,textfont=sf]{subfig}
\usepackage{textcomp}
\usepackage{stfloats}
\usepackage{url}
\usepackage{verbatim}
\usepackage{graphicx}
\usepackage{cite}
\usepackage[colorlinks]{hyperref}
\usepackage{multirow}
\usepackage{booktabs}
\usepackage{setspace}
\usepackage{makecell}
\usepackage[dvipsnames]{xcolor}
\usepackage{amssymb}

\begin{document}

\title{Defect Image Sample Generation With Diffusion Prior for Steel Surface Defect Recognition}

\author{Yichun Tai, Kun Yang, Tao Peng, Zhenzhen Huang, and Zhijiang Zhang
\thanks{The authors are with the School of Communication and Information Engineering, Shanghai University, Shanghai 200444, China (e-mail: taiyc@shu.edu.cn; 1987600116@shu.edu.cn; 17820125peng@shu.edu.cn; tristazz@shu.edu.cn; zjzhang@shu.edu.cn).}
\thanks{Corresponding author: Zhijiang Zhang.}}

\markboth{Journal of \LaTeX\ Class Files,~Vol.~14, No.~8, August~2021}%
{Shell \MakeLowercase{\textit{et al.}}: A Sample Article Using IEEEtran.cls for IEEE Journals}


\maketitle

\begin{abstract}
The task of steel surface defect recognition is an industrial problem with great industry values. 
The data insufficiency is the major challenge in training a robust defect recognition network.
Existing methods have investigated to enlarge the dataset by generating samples with generative models.
However, their generation quality is still limited by the insufficiency of defect image samples. 
To this end, we propose Stable Surface Defect Generation (StableSDG), which transfers the vast generation distribution embedded in Stable Diffusion model for steel surface defect image generation.
To tackle with the distinctive distribution gap between steel surface images and generated images of the diffusion model, we propose two processes. 
First, we align the distribution by adapting parameters of the diffusion model, adopted both in the token embedding space and network parameter space. Besides, in the generation process, we propose image-oriented generation rather than from pure Gaussian noises. 
We conduct extensive experiments on steel surface defect dataset, demonstrating state-of-the-art performance on generating high-quality samples and training recognition models, and both designed processes are significant for the performance. 

\textit{Note to Practitioners}---This article introduces StableSDG, a method that generates realistic defect images even with limited data. It overcomes the shortcomings of current deep learning approaches that need large datasets to train from scratch. Our solution is to adapt a text-to-image diffusion model for defect generation. The proposed strategy involves two processes: training to adapt token embeddings and model parameters, and generation from partially perturbed defect images. The results show enhanced generation quality and improved accuracy for recognition models trained on the expanded dataset. StableSDG can be practically applied to efficiently enlarge a defect dataset, even when starting with a small amount of data.

\end{abstract}


\section{Introduction}
\IEEEPARstart{S}{t}eel surface defect recognition aims at categorizing imperfections found on steel products. This practice plays a vital role in enhancing the quality of these products~\cite{ghorai2012automatic}. 
Traditional techniques, which require manual inspections for regular evaluations of structural and functional necessities, are not just time-consuming but also require significant manpower~\cite{hassan2019underground}. 

Contrarily, Automated Visual Inspection (AVI) provides distinct benefits regarding accuracy and effectiveness. Among existing methods for defect recognition in industrial manufacturing~\cite{ai2013surface,choi2014algorithm,dongyan2016defect}, deep learning techniques are increasingly prevalent~\cite{cheon2019convolutional,konovalenko2020steel,wang2021new}. These methods recognize defects with a neural network, which learns the pattern of defects from a collection of defect samples and their corresponding labels in an end-to-end process. Unfortunately, because defects do not occur on a predictable schedule, it often ends up with a shortage of images for network training. This can make it more difficult for the neural networks to work effectively.

To tackle this problem, a straightforward solution is to expand the defect dataset.
This can be done by creating more samples using generative models~\cite{yun2020automated,niu2020defect,zhao2023defect,zhang2022diversifying,zhang2021defect,duan2023few,li2023dls}.
Niu~\textit{et al.}~\cite{niu2020defect} propose surface defect-generation adversarial network (SDGAN) including two generators and four discriminators, to generate defect samples from defect-free images.
Zhang~\textit{et al.}~\cite{zhang2021defect} design Defect-GAN with the compositional layer-based architecture, to achieve the generation and removal of defects on surface images.
Inspired by the principle of image-to-image translation, Zhao~\textit{et al.}~\cite{zhao2023defect} design transP2P including transformer and U-Net, the former focuses on the global features perception, while the latter can better extract local detailed features, so as to transform defect-free images into defect images. However, training these generative models from scratch is challenging when the image samples are insufficient, which often leads to undesired patterns in the generated samples.

Recently, text-to-image generative models~\cite{ding2021cogview,ramesh2021zero,nichol2021glide,balaji2022ediffi,ramesh2022hierarchical,saharia2022photorealistic} have demonstrated impressive capabilities. They are embedded with vast image distribution, and can generate samples with high levels of fidelity and diversity. One such open-source model, Stable Diffusion~\cite{rombach2022high}, enables many powerful downstream applications through its efficient latent diffusion approach~\cite{wang2023exploiting,xia2023diffir,kawar2023imagic}. To further improve the generation fidelity to the content provided in a few reference images, existing methods have explored how to inject the shared concept in the reference images to the diffusion model for such customized generation~\cite{gal2022image,ruiz2023dreambooth,han2023svdiff,chen2023subject,chen2023disenbooth}. Chen~\textit{et al.}~\cite{chen2023disenbooth} first propose full-parameter adaptation to fit the diffusion model to the provided images.
In order to alleviate the catastrophic forgetting when the number of references is limited, Han~\textit{et al.}~\cite{han2023svdiff} propose to introduce limited trainable parameters to the diffusion model through SVG decomposition, such as to align with the provided images while avoiding over-fitting.
Chen~\textit{et al.}~\cite{chen2023subject} introduce an image-conditioned adapter to preserve the concept feature in the provided images without network parameter optimization. Despite that these methods can also generate high-fidelity samples with limited data resource, their generated content has a high intersection with the original generation distribution, which has a distinctive gap with the defect image distribution. As a result, using Stable Diffusion to directly generate steel surface defects is ineffective and could negatively impact classifier training by introducing low-quality data. In this paper, we propose StableSDG, which leverages the strong generative capabilities of Stable Diffusion model to generate defective image samples. To adapt the power of Stable Diffusion for generating high-quality steel surface defect data effectively, we propose the following pipeline that includes generator adaptation and data generation processes:

In the process of generator adaptation, rather than full-parameter adaptation, we use a combination of Textual Inversion~\cite{gal2022image} and low-rank adaptation~\cite{hu2021lora} to align the diffusion model with the distribution of defective images with limited parameter change. In the process of data generation, rather than generating from pure Gaussian noise, we propose to start the process from partially perturbed dataset samples. Our proposed pipeline is shown to be effective in generating defective image samples with limited data, achieving state-of-the-art performance in producing high-quality samples, and improving the performance of the defect recognition model.

In conclusion, our contributions are described as follows:
\begin{itemize}
\item To tackle the scarcity of defect image samples, we propose to employ the powerful Stable Diffusion model for steel surface defect generation. To our knowledge, this is the first time that text-to-image diffusion prior is employed for industrial image generation.

\item An effective pipeline StableSDG is proposed to adapt the text-to-image generative network for generating defect images with a large distribution gap. It composes of efficient adapting of network in both token embedding and network parameter space during training, and also a generation scheme from image-oriented initialization.

\item We conduct extensive experiments to demonstrate that StableSDG can generate defect images with higher fidelity than existing methods. Besides, it can effectively expand the defect dataset and substantially improve the accuracy by around 10\% on the task of continuous casting billet surface defect recognition.
\end{itemize}

The rest of this paper is organized as follows.  
In Section~\ref{Related Work}, we review the related work in defect image generation and recognition, and the existing effort that implementing Stable Diffusion prior to customized generation.
In Section~\ref{proposed method}, we present the strategies composed in StableSDG to generate defect images using the Stable Diffusion prior.
In Section~\ref{experiment}, we conduct experiments to evaluate the quality of the generated defect images, and validate the effectiveness for improving the performance of the recognition model.
Section~\ref{conclusion} summarizes our work.

\section{Related Work}
\label{Related Work}
In this section, we discuss the existing work related to defect image generation and recognition. 
We also cover the advancements in using the text-to-image diffusion prior for specified generation task provided with limited reference images.

\subsection{Defect Image Generation and Recognition}
The unpredictable occurrence of defects results in insufficient training data, making it very challenging to train a robust defect recognition model. To address this, existing methods can be categorized into two approaches: 1) developing algorithms to effectively train recognition models with limited data~\cite{song2022coarse,wang2022graph,wang2023few} and 2) generating additional samples to expand the dataset for training the recognition model~\cite{niu2020defect,zhao2023defect,zhang2022diversifying,zhang2021defect,duan2023few,li2023dls}.

To improve the performance through the training algorithm, Song~\textit{et al.}~\cite{song2022coarse} design a dynamic weighting module for discriminate features extraction, and a covariance metric module for similarity measurement.
Wang~\textit{et al.}~\cite{wang2023few} propose to pre-train the model with unlabeled data to learn effective image representation, then fine-tune with labeled data. In addition to recognizing the defect in the images, there are also methods designed to locate the defect in the images~\cite{dong2021defect,mo2020weighted,zhang2023duak}.

For defect dataset expansion, traditional approaches typically involve simulating defects through digital image processing or artificially introducing defects into defect-free workpieces~\cite{huang2009template,mery2005simulation,mery2002automated}. 
However, these methods can only create relatively straightforward defects with minimal diversity requirements, often resulting in significant wastage and increased costs.
Thanks to the excellent image generation performance of deep learning, data expansion becomes easily achievable by utilizing random variables sampled from known distributions.
The basic generative models involve Variational autoencoders (VAEs)~\cite{kingma2013auto} and Generative Adversarial Networks (GANs)~\cite{goodfellow2014generative}.
Then the models~\cite{li2018unsupervised,karras2021alias,sauer2022stylegan} derived from it, show satisfactory generative ability and have been widely used in general image generation tasks.
Based on these variations, several methods have been proposed for defect generation tasks. 
For example, Yun~\textit{et al.}~\cite{yun2020automated} propose a conditional CVAE, its input to the decoder is the encoding of the defect label concatenated with latent variable, so as to generate images for each type of defect, while GAN-based defect generation methods usually generate defect samples from defect-free images.
SDGAN~\cite{niu2020defect} contains two generators and four discriminators to expand the commutator cylinder surface defect image dataset by using a large number of defect-free images from industrial sites.
Zhang~\textit{et al.}~\cite{zhang2021defect} design Defect-GAN with the compositional layer-based architecture, to achieve the generation and removal of defects on surface images.
Inspired by the principle of image-to-image translation, Zhao~\textit{et al.}~\cite{zhao2023defect} introduce transP2P combining transformer and U-Net, the former focuses on the global features perception, while the latter can better extract local detailed features, so as to transform defect-free images into defect images.
Duan~\textit{et al.}~\cite{duan2023few} transfer the model pre-trained on defect-free images to the defect images to produce reasonable defect masks and accordingly manipulate the features within the masked regions.
Yang~\textit{et al.}~\cite{yang2024novel} train a denoising diffusion probabilistic model~\cite{ho2020denoising} from scratch to generate data for fault diagnosis.
Furthermore, some methods~\cite{zhang2022diversifying,li2023dls} are proposed to control regions and strength of generated defects. 

These methods all require training models from scratch with a vast collection of images. However, due to constraints in industrial settings, like intricate lighting conditions and noise interference, collecting a comprehensive set of defect-free or defect samples is difficult.
Motivated by the text-to-image diffusion model being embedded with a wide image distribution, we explore the use of text-to-image priors as a solution to defect image generation with limited available data.

\subsection{Customized Generation with Text-to-image Prior}
\begin{figure}[t] 
    \centering
    \includegraphics[scale=0.72]{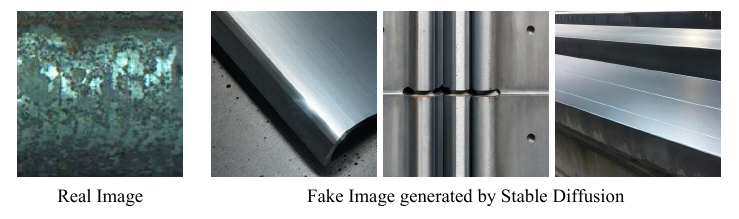}
    \caption{The images generated by Stable Diffusion model~\cite{rombach2022high} with the prompt "steel surface defect".}
    \label{SD_result}
\end{figure}

Text-to-image generators~\cite{nichol2021glide,balaji2022ediffi,ramesh2022hierarchical,saharia2022photorealistic} based on diffusion models~\cite{ho2020denoising,song2020denoising} show the high capacity in high-fidelity generation given diverse and abstract textual descriptions.
One of the most notable examples is Stable Diffusion model~\cite{rombach2022high}, which excels in generating images with low computation cost.
However, these generators cannot be tailored to individual preferences. 
Users are confined to the concepts the network has been trained on.
Considering the generation of defect images, when the term "steel surface defect" is inputted as a prompt, the images generated by the Stable Diffusion model miss the intricate textures and significantly differ from the steel surface defect images obtained from actual environments, as illustrated in Fig.~\ref{SD_result}.
To facilitate customized image generation, one could modify the Stable Diffusion model by incorporating new images into it. 
Yet, adapting the entire model with just a handful of images can significantly disrupt the learning process. The network may rapidly overfit to these new images, losing the broad array of concepts it was initially trained on. As a result, there is a need for a regulated adaptation method that enables the introduction of new concepts into the pre-trained model without compromising its original knowledge base.

A technique known as Textual Inversion~\cite{gal2022image} has been introduced, which focuses on learning a new token embedding using a small number of training examples and a prompt describing an unfamiliar concept. 
Zhang~\textit{et al.}~\cite{zhang2024compositional} introduce a spatial regularization approach aimed at balancing attention among composed concepts. Cai~\textit{et al.}~\cite{cai2024decoupled} address the interference of unrelated information by utilizing multiple tokens for image representation, mitigating its impact on the target concept. Additionally, Kumari~\textit{et al.}~\cite{kumari2023multi} achieve joint training for multiple concepts.
However, the effectiveness of these methods is constrained by the limited number of parameters that can be trained within the token embedding. Consequently, the generative capabilities of the model are not fully enhanced.
Another method for customizing Stable Diffusion, named DreamBooth~\cite{ruiz2023dreambooth}, aims to maintain the integrity of the original knowledge by retraining the model with a combination of original generated images and the new target images. 
The process of learning the entire generative model for each new concept introduced is not only expensive but also carries the risk of the model becoming too finely tuned to the new images. This overfitting can lead to negative consequences, such as catastrophic forgetting. In order to address these concerns, Chen~\textit{et al.}~\cite{chen2024subject} introduce apprenticeship learning to Text-to-Image generation, while the single apprentice model needs to be trained on a large amount of data. It should be noted that this method may not be suitable in situations where there is limited availability of data.
In terms of adapting large pre-trained models, low-rank adaptation~\cite{hu2021lora} is proposed to update a small set of parameters that can significantly influence the behavior of the model, avoiding the overfitting when adding new capabilities.
This approach allows for a targeted modification of the function of model while leveraging the rich representations learned during its initial extensive training phase.
Nevertheless, the content produced by these adaptive methods frequently shows substantial overlap with the original generative distribution, and high-quality generative results may not be achievable when the target image significantly deviates from the original generative distribution.

It should be noted that, to the best of our knowledge, this is the first instance where we have successfully preserved pre-existing knowledge while injecting new defect concepts into the Stable Diffusion model, even in the face of a distinctive distribution disparity between steel surface images and generated images of the diffusion model.
When faced with a scarcity of available images, our approach is capable of generating defect images with high fidelity. 
By using our model to expand the defect dataset, the performance of recognition models has seen a substantial enhancement.

\begin{figure}[t] 
    \centering
    \includegraphics[scale=0.72]{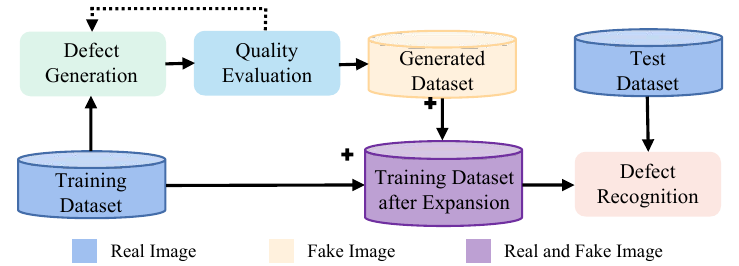}
    \caption{The overall pipeline, including defect generation, quality evaluation, and defect recognition. The dotted arrow indicates the quality evaluation is iteratively conducted until achieves the optimal hyperparameters, which are then used to construct generated dataset.}
    \label{pipeline}
\end{figure}

\section{Proposed Method}
\label{proposed method}
In this section, we introduce our method to implement the Stable Diffusion prior for expanding the dataset, under limited data resource. Specifically, we implement the Stable Diffusion model~\cite{rombach2022high} as the diffusion prior, which performs the diffusion in the latent space instead of image space, and is wildly used for image generation tasks~\cite{gal2022image,ruiz2023dreambooth,han2023svdiff,chen2023subject,chen2023disenbooth}. We conduct our proposed StableSDG, which is composed of two processes, for generating the images of each defect category. Through iterative quality evaluation, we tune hyperparameters to achieve optimal image generation. With the best hyperparameters, we generate high-quality images to expand the dataset. The generated images of each defect category along with the ground truth images are collected to train the defect recognition model. The overall pipeline is shown in Fig.~\ref{pipeline}.

\begin{figure*}[ht]
    \centering
    \includegraphics[scale=0.52]{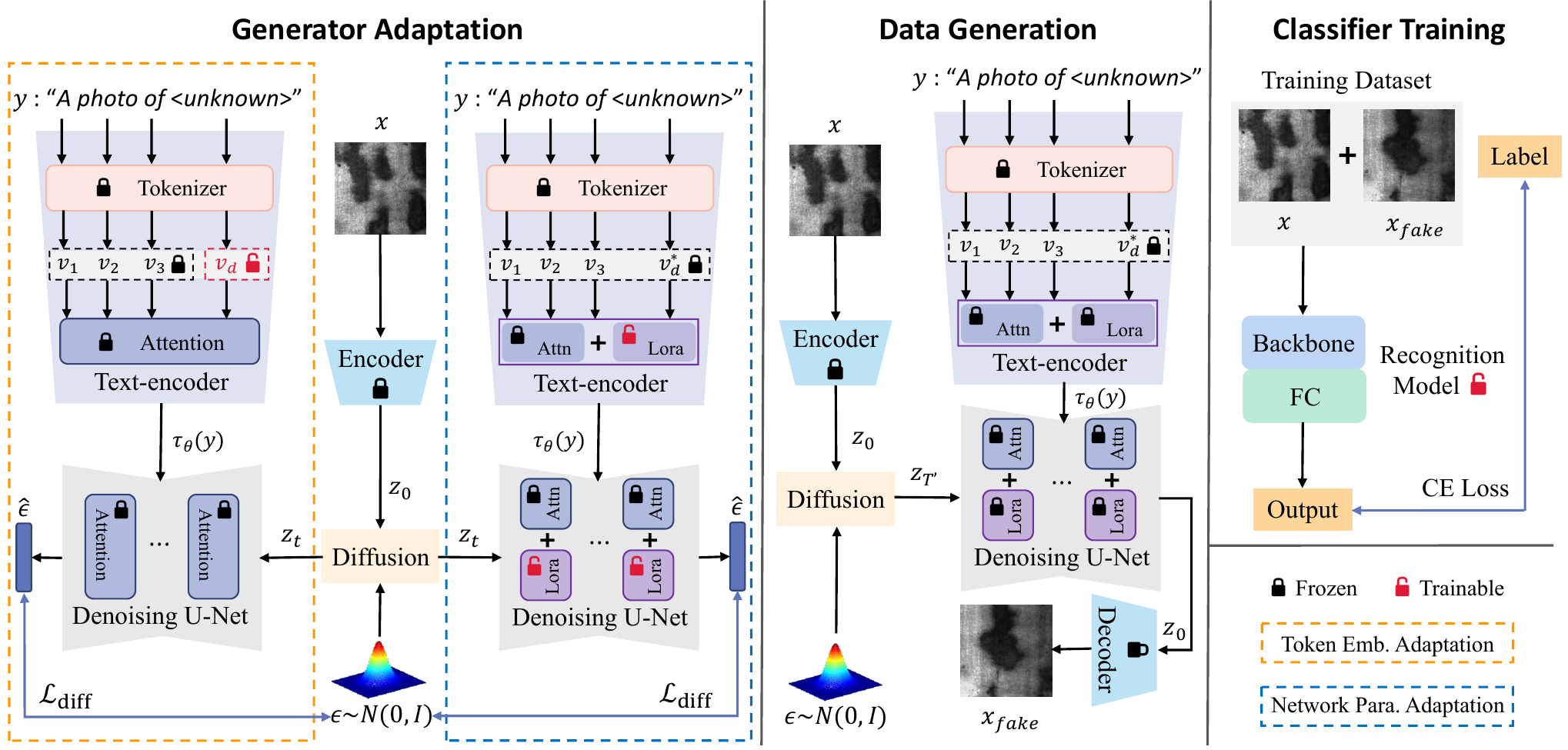}
    \caption{Overview of StableSDG. In the process of generator adaptation (Section~\ref{generator adaptation}), given the prompt $y$ (i.e., “\textit{A photo of} $<$\textit{unknown}$>$”) and the defect images $\mathbf{x}$ as input, we first optimize only the token embedding $v_d$ corresponding to the defect concept. Following this, we adapt the trainable matrices within the attention layers of both the text encoder and the U-net, using the previously optimized ${v}^{*}_d$. Next, we introduce image-oriented generation (Section~\ref{data generation}) to produce defect samples, which are then utilized to train a defect recognition classifier (Section~\ref{defect recognition}).}

    \label{StableSDG}
\end{figure*}

\subsection{Preliminary}
\label{sec:preliminary}

The diffusion model~\cite{ho2020denoising} is a class of generative models that learn the data distribution through a process of stepwise noise addition and subsequent recovery of the initial data. For the task of text-to-image generation, Stable Diffusion~\cite{rombach2022high} is wildly adopted, for its effective training and generation via conducting the diffusion and denoising process in the low-dimensional latent space. At its core, Stable Diffusion combines an autoencoder with a text-conditioned latent diffusion model. In this section, we will explore the key components of Stable Diffusion, from its basic autoencoder and latent diffusion techniques to the more advanced text-conditioned latent diffusion.

\subsubsection{Auto-encoder}
Stable Diffusion consists of an encoder and decoder to transform images to latent codes and vice versa. The encoder $E(\cdot)$ maps images $\mathbf{x} \in \mathbb{R}^D$ into latent codes $\mathbf{z}=E(\mathbf{x})$, where $\mathbf{z} \in \mathbb{R}^K$ and $K \ll D$. 
The decoder maps such latent codes back to images. With sufficient training, it holds that $D(E(\mathbf{x})) \approx \mathbf{x}$.

\subsubsection{Latent Diffusion Model}
The latent code $\mathbf{z} = E(\mathbf{x})$ is diffused into a series of increasingly noisy states ${\mathbf{z}_0 = \mathbf{z}, \mathbf{z}_1, ..., \mathbf{z}_T}$.
Each noisy state $\mathbf{z}_t$ follows the marginal distribution that $q(\mathbf{z}_t|\mathbf{z}) = \mathcal{N}\left(\alpha_t\mathbf{z}, \sigma_t^2 \boldsymbol{I}\right)$, where $\mathbf{z}_t$ can be sampled by $\mathbf{z}_t = \alpha_t \mathbf{z} + \sigma_t \epsilon, \epsilon \sim \mathcal{N}\left(\mathbf{0}, \boldsymbol{I}\right)$. Here $\alpha_t$ and $\sigma_t$ are scalars that $\sigma_t^2 + \alpha_t^2 = 1$. For the timesteps $t = 0, \cdots, T$, their $\alpha_t$ and $\sigma_t$ are configured such that as $t \rightarrow T$, the posterior distribution $q(\mathbf{z}_t|\mathbf{z})$ approaches a normal distribution. For the generation procedure, the model applies a reverse process to recover the clean latent $\mathbf{z}_0$ from the noisy end state $\mathbf{z}_T \sim \mathcal{N}\left(\mathbf{0}, \boldsymbol{I}\right)$. This reverse process can be represented by a Markov chain $p_\theta\left(\mathbf{z}_{0}\right)=p\left(\mathbf{z}_T\right) \prod_{t=1}^T p_\theta\left(\mathbf{z}_{t-1} \mid \mathbf{z}_t\right)$, which is a product of transition kernel that parameterized by $\theta$. Each transition kernel $p_\theta\left(\mathbf{z}_{t-1} | \mathbf{z}_t\right)$ is a normal distribution with mean $\mu_\theta\left(\mathbf{z}_t, t\right)$ and variance $\sigma_t^2 \boldsymbol{I}$. Estimating $\mu_\theta$ is equivalent to predicting the noise in $\mathbf{z}_t$, and such noise prediction is done with a neural network $ \epsilon_\theta(\cdot)$~\cite{ho2020denoising,song2020denoising}. In Stable Diffusion, $\epsilon_\theta(\cdot)$ is U-Net~\cite{ronneberger2015u} that is composed of convolution and attention operations. Besides, its noise prediction is under the text guidance, we next explain this in detail.

\subsubsection{Text-conditioned Latent Diffusion Model}
The noise prediction $\epsilon_\theta(\cdot)$ in Stable Diffusion is conditioned on the textual description, a.k.a. text prompt $y$. To utilize this condition, the text prompt is encoded using CLIP text encoder~\cite{radford2021learning}, which maps strings to low-dimensional token embeddings. We denote this encoder as $\tau_{\theta}(\cdot) = \tau^{r}_{\theta}(\tau^{n}_{\theta}(\cdot))$, it composes of a tokenizer $\tau^{n}_{\theta}(\cdot)$ followed by a Transformer network~\cite{vaswani2017attention} $\tau^{r}_{\theta}(\cdot)$, where we denote their pre-trained parameter sets as $\theta$.
In detail, the words in the string $y$ are tokenized into the token embeddings $v \in \mathbb{R}^{L \times C}$, via $v = \tau^{n}_{\theta}(y)$, where $L$ is the token embedding length and $C$ is the feature dimension. These token embeddings are then converted into a text embedding via $\tau^{r}_{\theta}(v) \in \mathbb{R}^{C}$.
The noise prediction is thus formulated as $\hat{\epsilon} = \epsilon_\theta(\mathbf{z}_t; \tau_{\theta}(y), t) = \epsilon_\theta(\mathbf{z}_t; \tau^{r}_{\theta}(\tau^{n}_{\theta}(y)), t)$, which takes the timestep $t$ and the encoded text prompt $\tau_{\theta}(y) = \tau^{r}_{\theta}(\tau^{n}_{\theta}(y))$ as conditions. With a dataset of text-image pairs $(\mathbf{x},y) \sim S$ and the pre-trained auto-encoder, the training objective for the network $\epsilon_\theta$ is the following loss function:
\begin{equation}
    \label{diffusion_training_objective}
\mathcal{L}_{\text{diff}} = \mathbb{E}_{t, \epsilon, \mathbf{x}, \mathbf{z} = E(\mathbf{x})}\left[\left\|\epsilon_\theta\left( \mathbf{z}_t; \tau^{r}_{\theta}(\tau^{n}_{\theta}(y)), t\right) - \epsilon\right\|_2^2\right],
\end{equation}
which takes the expectation of Mean Squared Error over the text-image pairs $(\mathbf{x},y) \sim S$, noise $
\epsilon \sim \mathcal{N}(\mathbf{0}, \boldsymbol{I})$, and timestep $t \sim \{1, \cdots, T\}$.
In terms of the conditional generation, on top of the transition kernel mentioned in the previous section, diffusion models balance the fidelity and diversity of this conditional generation using classifier-free guidance~\cite{ho2022classifier}:
\begin{equation}
    \label{classifier_free_guidance}
    \hat{\epsilon} = \epsilon_\theta\!\left(\mathbf{z}_t;t\right)+\omega_{cfg}\left[ \epsilon_\theta\left(\mathbf{z}_t;\tau^{r}_{\theta}(\tau^{n}_{\theta}(y)), t\right)-\epsilon_\theta\left(\mathbf{z}_t;t\right) \right],
\end{equation}
where $\epsilon_\theta\left(\mathbf{z}_t ; t\right)$ is the prediction of noise without text guidance, and $\omega_{cfg}$ is a scalar that adjusts the influence of the condition on the generative process. Generating images in Stable Diffusion is concluded as iterating $\mathbf{z}_{t-1} = \alpha_{t-1} \frac{\mathbf{z}_t - \sigma_t \hat{\epsilon}}{\alpha_t} + \sigma_t \epsilon$~\cite{ho2020denoising}, which starts from pure Gaussian $\mathbf{z}_T$ and ends in $\mathbf{z}_0$ that can be decoded to image via $\mathbf{x} = D(\mathbf{z}_0)$.

\begin{algorithm}[!t]
\caption{StableSDG}\label{alg_StableSDG}
\begin{algorithmic}
\STATE \textbf{Input}: prompt $y$, defect images of the single category $p(\mathbf{x})$, Stable Diffusion $\epsilon_\theta(\mathbf{z}_t; \tau^{r}_{\theta}(\tau^{n}_{\theta}(y)), t)$,  guidance scale $\omega_{cfg}$, strength $s$ \\
\STATE {\textbf{\textsc{Generator Adaptation}}}\\
\STATE Initialize $v = \tau^{n}_{\theta}(y)= [v', v_d]$ \\
\STATE \textit{ // Token Embedding Adaptation} \\
\STATE While not converge, do: 
\STATE \hspace{0.5cm} Sample $\mathbf{\mathbf{x}} \sim p(\mathbf{\mathbf{x}})$, $\epsilon \sim \mathcal{N}\left(\mathbf{0},\boldsymbol{I}\right)$, $t \in \{1, \dots, T\}$
\STATE \hspace{0.5cm} $\mathbf{z}_0 = \emph{E}(\mathbf{x})$
\STATE \hspace{0.5cm} $\mathbf{z}_t = \alpha_t \mathbf{z}_0 + \sigma_t \epsilon$
\STATE \hspace{0.5cm} Update $v_d$ with $\nabla_{v_d} \mathcal{L} = \|\epsilon_\theta(\mathbf{z}_t; \tau^{r}_{\theta}(v), t)-\epsilon\|^2_2$
\STATE  Denote ${v}^{*}_d$ as the optimized token embedding \\
\STATE $v^* = [v', v^*_d]$ \\
\STATE \textit{ // Network Parameter Adaptation} \\
\STATE While not converge, do:
\STATE \hspace{0.5cm} Sample $\mathbf{\mathbf{x}} \sim p(\mathbf{\mathbf{x}})$, $\epsilon \sim \mathcal{N}\left(\mathbf{0},\boldsymbol{I}\right)$, $t \in \{1, \dots, T\}$
\STATE \hspace{0.5cm} $\mathbf{z}_0 = \emph{E}(\mathbf{x})$
\STATE \hspace{0.5cm} $\mathbf{z}_t = \alpha_t \mathbf{z}_0 + \sigma_t \epsilon$
\STATE \hspace{0.5cm} Update $\theta$ with $\nabla_{\theta} \mathcal{L}=\|\epsilon_\theta(\mathbf{z}_t;\tau^{r}_{\theta}(v^*),t)-\epsilon\|^2_2$\\
\STATE  Denote ${\theta}^{*}$ as the optimized network parameter \\
\STATE {\textbf{\textsc{Data Generation}}}\\
\STATE \textit{ // Image-oriented Generation} \\
\STATE Sample $\mathbf{\mathbf{x}} \sim p(\mathbf{\mathbf{x}})$, $\epsilon \sim \mathcal{N}\left(\mathbf{0},\boldsymbol{I}\right)$ \\
\STATE $T^{'}=sT$ \\
\STATE $\mathbf{z}_{T^{'}}=\alpha_{T^{'}}E(\mathbf{x})+\sigma_{T^{'}} \epsilon$ \\
\FOR{$i=T^{'}$ to $1$} 
\STATE $\hat{\epsilon}=\epsilon_{\theta^*}(\mathbf{z}_t ; t)+\omega_{c f g}\left[\epsilon_{\theta^*}(\mathbf{z}_t ; \tau_{\theta^*}^r\left(v^{*}\right), t)-\epsilon_{\theta^*}(\mathbf{z}_t ; t)\right]$
\STATE $\epsilon \sim \mathcal{N}\left(\mathbf{0},\boldsymbol{I}\right)$
\STATE $\mathbf{z}_{i-1} = \alpha_{i-1} \frac{\mathbf{z}_i - \sigma_i \hat{\epsilon}}{\alpha_i} + \sigma_i \epsilon$
\ENDFOR 
\STATE Output $\mathbf{x} = D(\mathbf{z}_0)$\\
\end{algorithmic}
\label{alg1}
\end{algorithm}

\subsection{StableSDG}
\label{sec:defect generation}
To adapt the power full Stable Diffusion for generating high-quality steel surface defect data, we propose StableSDG, which includes two processes, i.e. generator adaptation and data generation.
Fig.~\ref{StableSDG} represents the overview of our method, and the details are shown in Algorithm~\ref{alg_StableSDG}. Next we present the two processes in detail.

\subsubsection{Generator Adaptation}
\label{generator adaptation}
This process adapts Stable Diffusion to generate images of steel surface defects, which deviates significantly from the model's original image distribution. We propose the following two strategies in this process.

\textbf{Token Embedding Adaptation.}
The discrepancy between the defect images from production collection and the images generated by the Stable Diffusion model, as indicated in Fig.~\ref{SD_result}, partially stems from an inaccurate text prompt. This can be enhanced by having the text prompt $y$ better represent steel surface defect images. We adopt the strategy proposed by~\cite{gal2022image}, which optimizes the token embeddings $v=\tau^{n}_{\theta}(y)$ as they are differentiable. 
Specifically, consider $y$ is written as \textit{A photo of $<$unknown$>$}, its $v$ is a sequence of token embeddings, i.e. $v = \left[v_1, \cdots, v_L\right], v_1, \cdots, v_L \in \mathbb{R}^C$, we use the notation $v_d$ to index the part of embeddings in this sequence that correspond to the sub-string of \textit{$<$unknown$>$}, which refers to the specific defect concept.
In this case, we can denote $v'$ as the sequence of other token embeddings and $v$ as $\left[v', v_d\right]$. We optimize $v_d$ via:
\begin{equation}
    \label{textural inversion}
    {v}^*_d\!=\!\underset{v_d}{\operatorname{argmin}}\mathbb{E}_{t, \epsilon, \mathbf{x}, \mathbf{z} = E(\mathbf{x}), v = [v', v_d]}\!\left[\!\left\|\epsilon_\theta\!\left(\!\mathbf{z}_t;\!\tau^{r}_{\theta}(v),\!t\!\right)\!-\!\epsilon\right\|_2^2\!\right]\!,
\end{equation}
which minimizes the training loss as in Equation~\ref{diffusion_training_objective} and takes the expectation over a specific category of defect images $\mathbf{x}$ (e.g. crazing, inclusion), diffusion timesteps $t$ and noise $\epsilon$. We denote the full token embeddings after the optimization as $v^* = [v', {v}^{*}_d]$, which is used for the following adaptation.

\textbf{Network Parameter Adaptation.} 
Given that optimization within the token embedding space alone may bring with limited improvement on the fidelity of the generated content, further enhancement can be achieved through adaptation on the network parameters. However, since the Stable Diffusion model comprises billions of parameters, compared with the limited amount of defect images, attempting to fine-tune the entire network parameters, as done in~\cite{ruiz2023dreambooth}, would lead to over-fitting. We fine-tune the model through low-rank adaptation~\cite{hu2021lora}, which allows for constrained parameter change. For the weight of each dense layer $W_0 \in \mathbb{R}^{d \times k}$, it conducts parameter adaptation by imposing a low-rank decomposition:
\begin{equation}
W_0+\Delta W=W_0+B A,
\label{lora}
\end{equation}
where $\Delta W$ is the change in weights, and is represented by the product of two matrices, $B \in \mathbb{R}^{d \times r}$ and $A \in \mathbb{R}^{r \times k}$. Here, the rank $r$ is significantly smaller than the minimum of $d$ and $k$. As a result, the number of parameters for adaptation is significantly reduced from $d \times k$ to $(d + k) \times r$. 
In practice, we set $r=1$.
Matrix $A$ is initialized following a Gaussian distribution, and matrix $B$ is initialized to zero, which ensures that $\Delta W = BA = 0$ at the start of training. Throughout the training process, $A$ and $B$ are adjusted while $W_0$ remains static. As shown in Fig.~\ref{StableSDG}, such low-rank adaptation is conducted for all the attention layers~\cite{vaswani2017attention} in the CLIP text encoder $\tau^{r}_{\theta}(\cdot)$ and the U-net $\epsilon_\theta(\cdot)$, by adapting their original parameter sets from $\theta$ to $\theta^*$. Both $\epsilon_\theta(\cdot)$ and $\tau^{r}_{\theta}(\cdot)$ are fine-tuned using the following objective function:
\begin{equation}
\label{lora_theta}
\theta^* = \underset{\theta}{\operatorname{argmin}} \mathbb{E}_{t, \epsilon, \mathbf{x}, \mathbf{z} = E(\mathbf{x})}\left[\left\|\epsilon_\theta\left(\mathbf{z}_t; \tau^{r}_{\theta}(v^*), t\right) - \epsilon\right\|_2^2\right].
\end{equation}
This approach facilitates adaptation of the model with a much lower risk of over-fitting, because of the drastically reduced parameter space.

\subsubsection{Data Generation}
\label{data generation}
To enhance the quality of defect image generation with the adapted Stable Diffusion, we introduce the following strategy.

\textbf{Image-oriented Generation}. 
Based on the token embeddings $v^*$ and the network parameter $\theta^*$, we generate the latent code $\mathbf{z}_0$ from half-perturbed code $\mathbf{z}_{T^{'}}=\alpha_{T^{'}} E(\mathbf{x})+\sigma_{T^{'}} \epsilon$, where $T^{'}=sT$ denotes the decreased maximum degree of noise diffusion, and $s \in (0,1)$ is the denoising strength. The scalar $s$ controls the similarity between the generated image $D(\mathbf{z}_0)$ and the original image $\mathbf{x}$. The smaller the value, the higher the similarity.
The image-oriented generation can be represented as $p_{\theta^*}\left(\mathbf{z}_{0}\right)=p\left(\mathbf{z}_{T^{'}}\right) \prod_{t=1}^{T^{'}} p_{\theta^*}\left(\mathbf{z}_{t-1} \mid \mathbf{z}_t\right)$. With $\mathbf{z_0} \sim p_{\theta^*}(\mathbf{z}_0)$, the generated image is $D(\mathbf{z}_0)$. 
\begin{table}[t]
	\setlength{\tabcolsep}{0.2cm}
    \renewcommand\arraystretch{1.2}
	\begin{center}
	\caption{
     The hyperparameters of image-oriented generation. $\omega_{cfg}$ and $s$ are scaling factor and denoising strength, respectively.
     The defect categories are detailed in~\ref{experimental_settings}.
	}
	\label{parameter_setting}
    \footnotesize
	\begin{tabular}{c|cccccc|cccc}
	\hline\hline
 {\multirow{2}*{}}&\multicolumn{6}{c|}{NEU}&\multicolumn{4}{c}{CCBSD}\\    
    &Cr&In&Pa&PS&RS&Sc&Inc&Ind&Ox&SG\\
    \hline
$s$&0.2&0.9&0.4&0.5&0.1&0.5&0.5&0.2&0.6&0.5\\
$\omega_{cfg}$&2&9&6&3&8&7&5&7&3&4\\
\hline\hline
\end{tabular}
\end{center}
\end{table}

\subsection{Quality Evaluation}
To improve the quality of generated defect images, we iteratively adjust the model hyperparameters based on the Fréchet Inception Distance (FID) evaluation metric\cite{heusel2017gans}.
FID measures the similarity between real and generated image distributions through \begin{equation}
\mathrm{FID}={\left\|\mu_r-\mu_g\right\|}^2+\operatorname{Tr}\left(C_r+C_g-2\left(C_r C_g\right)\right)^{1 / 2},
\label{FID}
\end{equation}
where $\mu_r$ and $\mu_g$ are the mean feature vectors for the real and generated images, respectively. $C_r$ and $C_g$ are the covariance matrices for the feature vectors of the real and generated images. 
We adjust the guidance scale $\omega_{cfg}$ and strength $s$ of each defect category for lower FID scores.
The final hyperparameters of image-oriented generation are detailed in Table~\ref{parameter_setting}.

\subsection{Defect Recognition}
\label{defect recognition}
After the quality evaluation, we can obtain the optimal distribution of the generated defect dataset $p_{\theta^*}(\mathbf{x})$.
And then, recognition models~\cite{krizhevsky2012imagenet,simonyan2014very,he2016deep,iandola2016squeezenet,huang2017densely} are trained on the expanded defect dataset.
The size of the input defect image is 200 $\times$ 200 $\times$ 3.
The training objective for the recognition network $R(\cdot)$ is to minimize the cross-entropy loss function:
\begin{equation}
    \label{recognition_training_objective}
\mathcal{L}_{\text{cls}} =\mathbb{E}_{p(c \mid \mathbf{x})}[-\log \hat{p}(c \mid \mathbf{x})],
\end{equation}
where $p\left(c \mid \mathbf{x}\right)$ is the empirical distribution of the training sets, and $\hat{p}\left(c \mid \mathbf{x}\right)$ is the predicted distribution from the $R(\cdot)$.
The image $\mathbf{x}$ is sampled from the combination of the original defect dataset and generated defect images.

\begin{figure}[t] 
    \centering
    \includegraphics[scale=0.6]{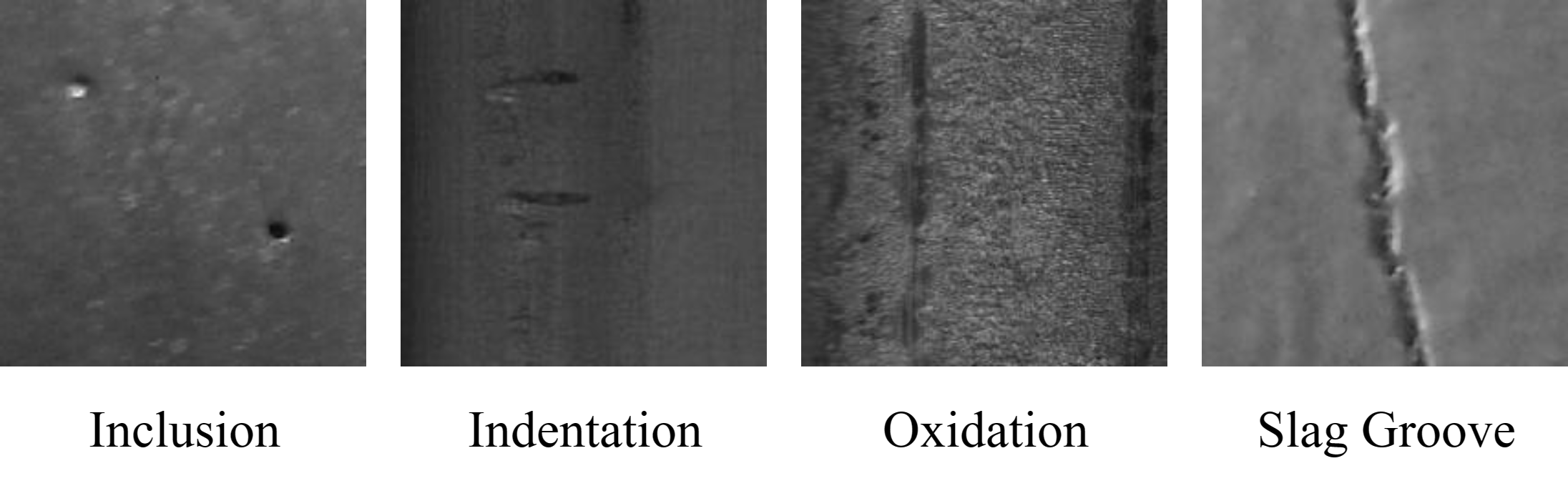}
    \caption{The illustration of defect categories in CCBSD.}
    \label{CCBSD}
\end{figure}

\section{Experiment}
\label{experiment}
In this section, we first introduce the experimental settings (\ref{experimental_settings}), and then validate the effectiveness of StableSDG with the ablation study (\ref{ablatin_study}).
To verify the superiority of the proposed method, we evaluate the generated image quality and the performance of the recognition models on the Northeastern University surface defect database~\cite{song2013noise} (\ref{performance_NEU}), and the continuous casting billet surface defect dataset (\ref{performance_CCBSD}) respectively.

\subsection{Experimental Settings}
\label{experimental_settings}

\textbf{Datasets}. We conduct experiments on the Northeastern University surface defect database (NEU)~\cite{song2013noise}, an open-source steel surface defect dataset. It consists of six typical surface defects of hot-rolled steel strip, including Crazing (Cr), Inclusion (In), Patches (Pa), Pitted Surface (PS), Rolled-in Scale (RS) and Scratches (Sc). 
Each category has 300 grayscale defect samples with 200 × 200 resolution.
To verify the performance of the proposed method in practical application, we build the continuous casting billet surface defect dataset (CCBSD). 
Following the prior art on the process the steel images from industrial production~\cite{song2013noise}, we convert them to grayscale followed by binarization, so as to get the region of interest, i.e., the region where the continuous casting billet is located.
Subsequently, the image is segmented into multiple sub-images with a 1:1 aspect ratio, each being resized to a resolution of 200 $\times$ 200 pixels.
In order to minimize the possibility of misrecognition, we adopt human annotators to generate the ground truth labels.
Due to the insufficiency of defect images, the constructed initial dataset only contains 200 samples per class.
The dataset will be made publicly available.
Fig.~\ref{CCBSD} shows common surface defects of continuous casting billet such as inclusion (Inc), indentation (Ind), oxidation (Ox) and slag groove (SG).

\begin{table}[t]
	\setlength{\tabcolsep}{0.26cm}
    \renewcommand\arraystretch{1.2}
	\begin{center}
	\caption{
	FID scores of generated images with different stages of proposed method shown in Algorithm~\ref{alg_StableSDG}.
	}
	\label{ablatin_study_table}
    \footnotesize
	\begin{tabular}{cccc}
	\hline\hline
    \multicolumn{2}{c}{Generator Adaptation}&{Data Generation}&\multirow{2}*{FID Scores $\downarrow$}\\
    \cmidrule{1-3}
    Token Emb.&Network Para.&Image-oriented&\\    
    \hline
    \checkmark&&&245.35\\
    \checkmark&&\checkmark&138.19\\
    &\checkmark&&110.38\\
    &\checkmark&\checkmark&70.56\\
    \checkmark&\checkmark&&106.52\\
    \checkmark&\checkmark&\checkmark&\textbf{64.49}\\
    \bottomrule[1pt]
    \multicolumn{2}{c}{\multirow{2}*{Full-parameter adaptation}}&&111.61\\
    &&\checkmark&70.96\\
\hline\hline
\end{tabular}
\end{center}
\end{table}

\begin{figure}[!t] 
    \centering
    \includegraphics[scale=0.54]{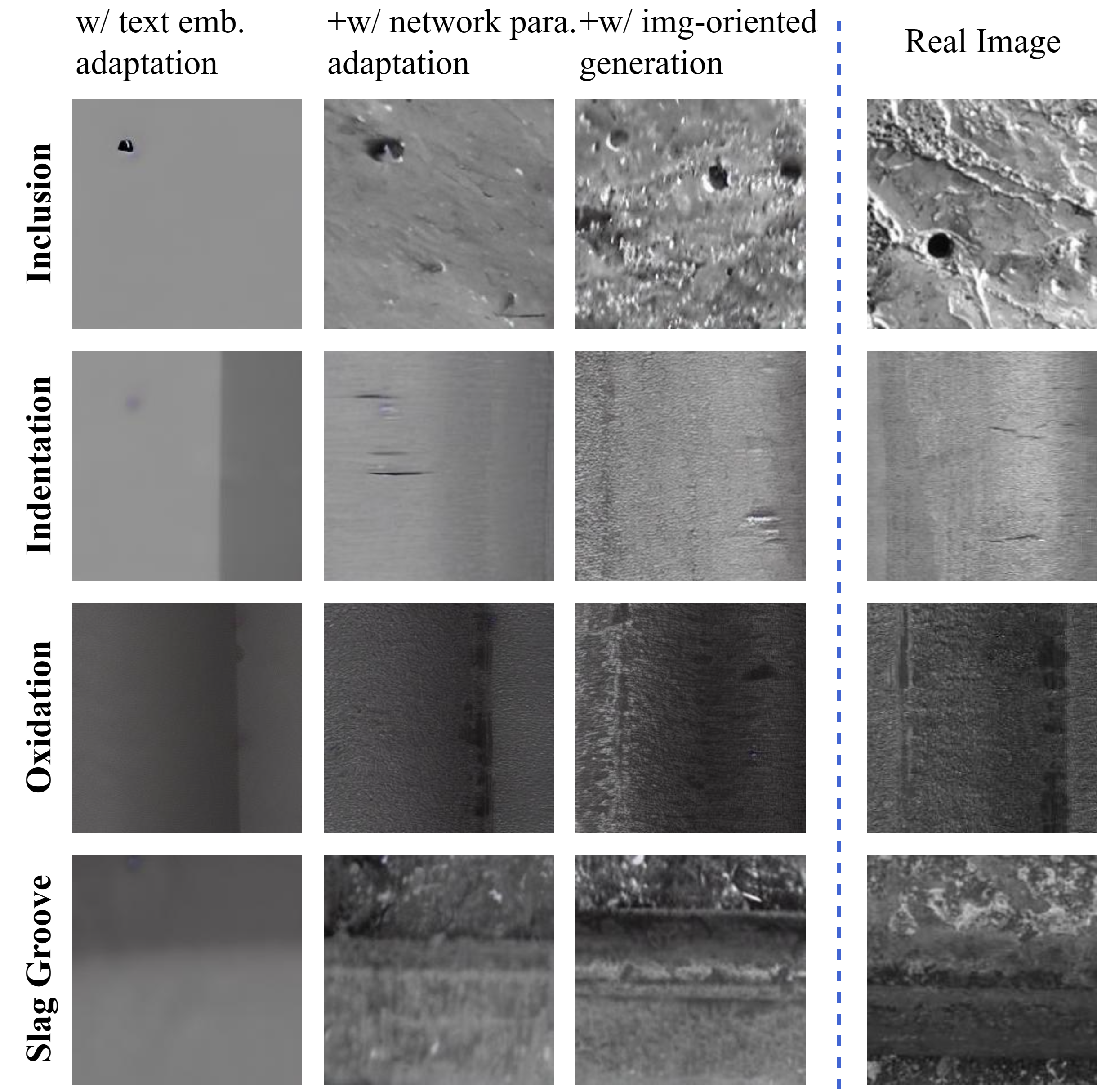}
    \caption{Intermediate results of StableSDG for various defect categories.}
    \label{fig_abla}
\end{figure}

\textbf{Implementations}. 
For data generation, we set up benchmarks on NEU and CCBSD datasets, and compare with existing state-of-the-art methods, i.e. DCGAN~\cite{li2018unsupervised}, StyleGAN3~\cite{karras2021alias}, DDIM~\cite{song2020denoising}, Textual Inversion~\cite{gal2022image}, DreamBooth~\cite{ruiz2023dreambooth}, and Yang~\textit{et al.}~\cite{yang2024novel}.
Our StableSDG is based on Stable Diffusion v1.5~\cite{stable-diffusion-v1-5}. We conduct training with Adam optimizer~\cite{kingma2014adam} and the batch size of 4.
Both the generator adaptation and the data generation processes run for 1,000 iterations. The learning rates for both stages are 5e-4 and 1e-4, respectively. For the baseline methods, we adhere to their official implementations and tune the hyperparameters to ensure the best possible performance.
Each model is utilized to generate 1,000 images per defect category. 
For classifier training, we use data augmentation, i.e. including rotating within $\left[-10^{\circ}, +10^{\circ}\right]$ and random flipping along the horizontal and vertical axes. 
We conduct a thorough experiment with various classifiers, i.e. VGG~\cite{simonyan2014very}, ResNet~\cite{he2016deep}, SqueezeNet~\cite{iandola2016squeezenet} and DenseNet~\cite{huang2017densely}. We train these networks with Adam optimizer, the learning rate of 1e-4, batch size of 32, select the best-performing networks based on their validation performance and evaluate on the test sets.
All experiments are conducted with 1 NVIDIA V100 GPU.

\begin{figure}[!t] 
    \centering
    \includegraphics[scale=0.42]{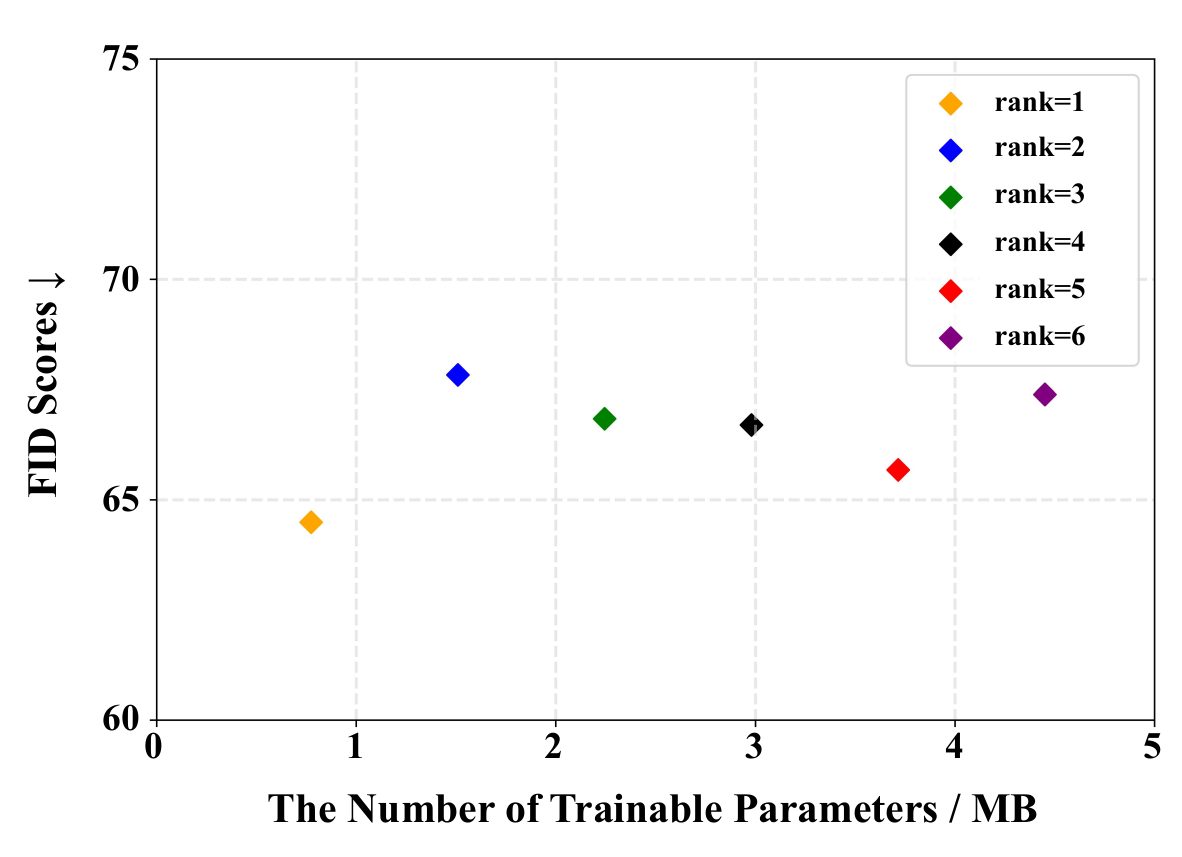}
    \caption{FID scores and trainable parameters across various LoRA ranks $r$.}
    \label{fig_rank_NEU}
\end{figure}

\begin{table}[!t]
	\setlength{\tabcolsep}{0.8cm}
    \renewcommand\arraystretch{1.2}
	\begin{center}
	\caption{
    FID scores of generated images with different prompt $y$. Take the defect category "Pa" in NEU as an example.}
	\label{ablation_prompt}
    \footnotesize
	\begin{tabular}{c|c}
	\hline\hline
	Prompt $y$&FID Scores $\downarrow$\\    
    \hline
"\textit{A photo of defect}"& {65.56} \\
"\textit{A photo of patches}"& {68.17}\\
"\textit{A photo of $<$unknown$>$}"& {\textbf{64.60}}\\
\hline\hline
\end{tabular}
\end{center}
\end{table}

\begin{table}[!t]
	\setlength{\tabcolsep}{0.24cm}
    \renewcommand\arraystretch{1.2}
	\begin{center}
	\caption{
    FID scores of generated images with different guidance scale $\omega_{cfg}$ and strength $s$.
    Take the defect category "Pa" in NEU as an example.
	}
	\label{parameter_setting_pa}
    \footnotesize
	\begin{tabular}{cc|ccccc}
	\hline\hline
	\multicolumn{2}{c|}{\multirow{2}*{}}&\multicolumn{5}{c}{Guidance Scale $\omega_{cfg}$}\\    
    &&3&4&5&6&7\\
    \hline
    \multirow{5}*{Strength $s$}&0.3&112.02&103.54&97.42&94.09&94.56\\
    &0.4&97.95&93.12&90.73&\textbf{89.69}&90.51\\
    &0.5&99.93&94.31&93.34&94.92&97.68\\
    &0.6&99.60&96.88&94.83&97.71&97.20\\
    &0.7&99.56&95.02&93.28&90.04&95.87\\
\hline\hline
\end{tabular}
\end{center}
\end{table}

\begin{figure*}[!t]
    \centering
    \includegraphics[scale=0.68]{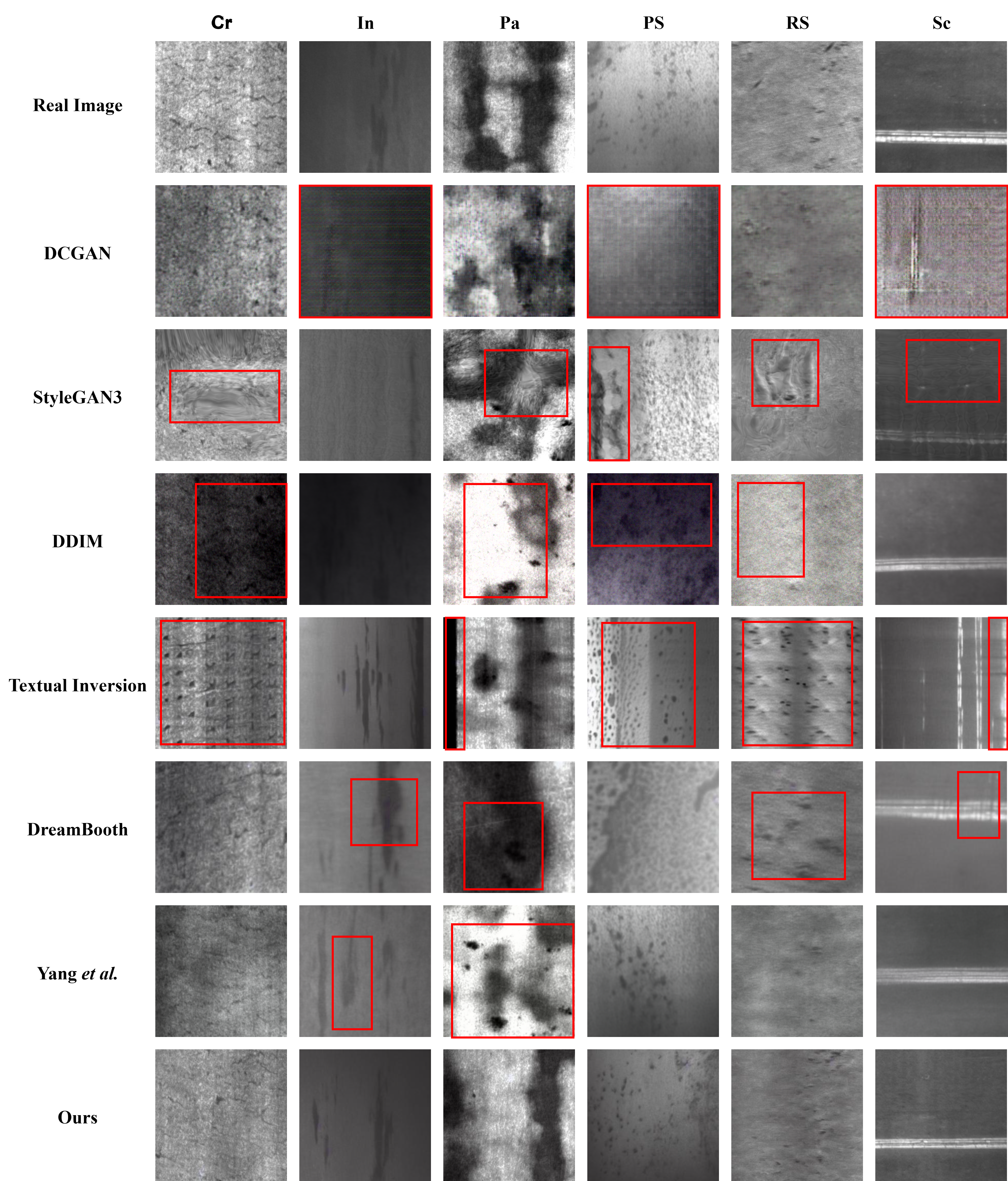}
    \caption{Qualitative comparison of StableSDG with other models on NEU. Artifacts are highlighted in red boxes, with details in Section~\ref{performance_NEU}.}
    \label{visualized_result_on_NEU}
\end{figure*}

\subsection{Ablation Study}
\label{ablatin_study}
We conduct ablation studies on NEU to observe the impact of different components and identify the best configuration and hyperparameters for our proposed method.

\textbf{Impact of Multiple Stages}. To assess the efficacy of our proposed method with the three stages, i.e. token embedding adaption, network parameter adaptation and image-oriented generation, we conduct an evaluation of the quality of images generated with these stages.
From Table~\ref{ablatin_study_table}, we can find that, 1) omitting any stage of our proposed StableSDG leads to a deterioration in image quality, thereby confirming the importance of each stage in the process; 2) compared with full-parameter adaptation, i.e., re-training the Stable Diffusion model on the limited defect image collection~\cite{ruiz2023dreambooth}, our method yields improved generation quality. This improvement stems from our effective adaptation to the data distribution of steel surface defects, reducing the deterioration usually caused by full-parameter fine-tuning and catastrophic forgetting. 
We also present intermediate results of StableSDG through its three stages and across various defect categories in CCBSD, as illustrated in Fig.~\ref{fig_abla}. The model incrementally generates samples that closely resemble real defect images.

\textbf{Impact of the Prompt}. Different text prompts $y$ within the text-to-image diffusion prior might influence generation quality. We explore several text prompts and quantitatively evaluate their effects in Table~\ref{ablation_prompt}. For each prompt, we optimize the token embedding related to the defect category, such as \textit{defect}, \textit{patches}, and \textit{$<$unknown$>$}, during the token embedding adaptation stage. The results show that using \textit{$<$unknown$>$} for new defect categories results in lower FID scores, suggesting that \textit{$<$unknown$>$} as an initialization helps avoid local optima and performs better than more specific prompts like \textit{defect}, which yield poorer outcomes.

\begin{table}[!t]
    \footnotesize
	\setlength{\tabcolsep}{0.04cm}
    \renewcommand\arraystretch{1.2}
	\begin{center}
	\caption{
    Quantitative comparison among various image generative models trained on NEU.
	}
	\label{quality_NEU}
	\begin{tabular}{c|cccccc}
	\hline\hline
	\multirow{2}*{}&\multicolumn{6}{c}{FID Scores $\downarrow$}\\
    &Cr&In&Pa&PS&RS&Sc\\
    \hline
{DCGAN \textcolor{gray}{\scriptsize{(CCC~18~\cite{li2018unsupervised})}}}&169.83&230.78&153.15&300.34&182.07&291.41\\
{StyleGAN3 \textcolor{gray}{\scriptsize{(NIPS~21~\cite{karras2021alias})}}}&97.12&209.54&189.30&148.97&111.57&221.25\\
{DDIM \textcolor{gray}{\scriptsize{(ICLR~21~\cite{song2020denoising})}}}&151.51&83.26&108.71&139.12&216.07&105.43\\
{{Textual Inversion~\textcolor{gray}{\scriptsize{(ICLR~23~\cite{gal2022image})}}}}&{246.53}&{166.09}&{173.13}&{139.13}&{278.76}&{151.64}\\
{{DreamBooth~\textcolor{gray}{\scriptsize{(CVPR~23~\cite{ruiz2023dreambooth})}}}}&{79.64}&{173.97}&{133.84}&{116.04}&{136.93}&{119.49}\\
{{Yang~\textit{et al.}~\textcolor{gray}{\scriptsize{(TII~24~\cite{yang2024novel})}}}}&{68.54}&{78.07}&{91.86}&{75.22}&{82.41}&{96.99}\\
{StableSDG (Ours)}&\textbf{46.90}&\textbf{72.26}&\textbf{64.60}&\textbf{59.33}&\textbf{58.01}&\textbf{85.84}\\
\hline\hline
\end{tabular}
\end{center}
\end{table}

\textbf{Impact of the LoRA Rank}. 
Figure~\ref{fig_rank_NEU} shows how the LoRA rank $r$ influences both the generation quality and the count of trainable parameters. While increasing $r$ leads to a higher number of trainable parameters, it does not significantly improve performance. Consequently, we have chosen to set $r$ to 1 in our study.

\textbf{Impact of Guidance Scale and Strength}. Additionally, we underscore the importance of the guidance scale $\omega_{cfg}$ and strength $s$ as the hyper-parameters for image-oriented generation.
As detailed in Table~\ref{parameter_setting_pa}, we can see about 20\% performance enhancement between the highest and lowest metrics, indicating that good hyperparameters bring significant improvement in the quality of generated images.

\begin{figure}[!t] 
    \centering
    \includegraphics[scale=0.42]{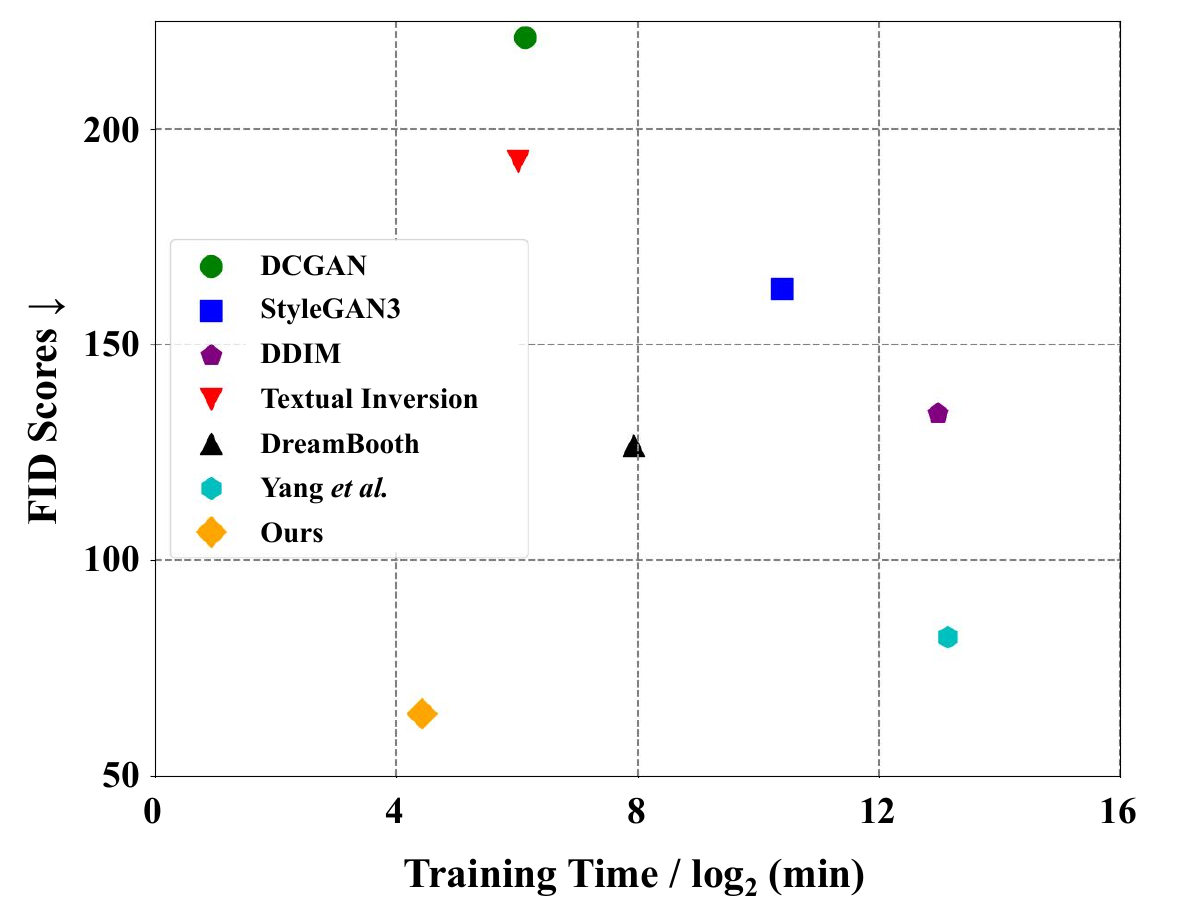}
    \caption{Comparison of generative capacity and training efficiency of StableSDG with other image generative models on NEU.}
    \label{fig_speed_NEU}
\end{figure}

\subsection{Performance on NEU}
\label{performance_NEU}

\textbf{Generation quality}. We evaluate StableSDG for defect image generation on NEU dataset.
Table~\ref{quality_NEU} shows the image generation performance, as measured by FID, in six categories of the dataset.
The qualitative comparisons are also shown in Fig.~\ref{visualized_result_on_NEU}.
Regions with abnormal textures or anomalous patterns in the generated images are marked in red boxes.
It is observed that images generated by DCGAN contain some abnormal textures within the backgrounds, notably within the "In", "PS" and "Sc" defect categories.
The outputs from StyleGAN3 closely resemble the original images overall. However, there are noticeable anomalous patterns in specific regions across most of the defect categories.
The results of Textual Inversion exhibit distinct bright and dark stripes, whereas the backgrounds of images generated by DreamBooth are more uniform, though they still have areas that are not entirely satisfactory. Regarding DDIM, the images it generates are somewhat lacking in detail, and there is a noticeable inconsistency in brightness, particularly in the images from the defect category "Pa", as is the case with Yang \textit{et al.} In contrast, our StableSDG achieves the lowest FID in each defect category compared with other methods. Additionally, our method incurs the lowest training cost, as illustrated in Fig.~\ref{fig_speed_NEU}, showing it requires the least amount of training time (21.60 minutes).

\begin{figure}[!t] 
    \centering
    \includegraphics[scale=0.42]{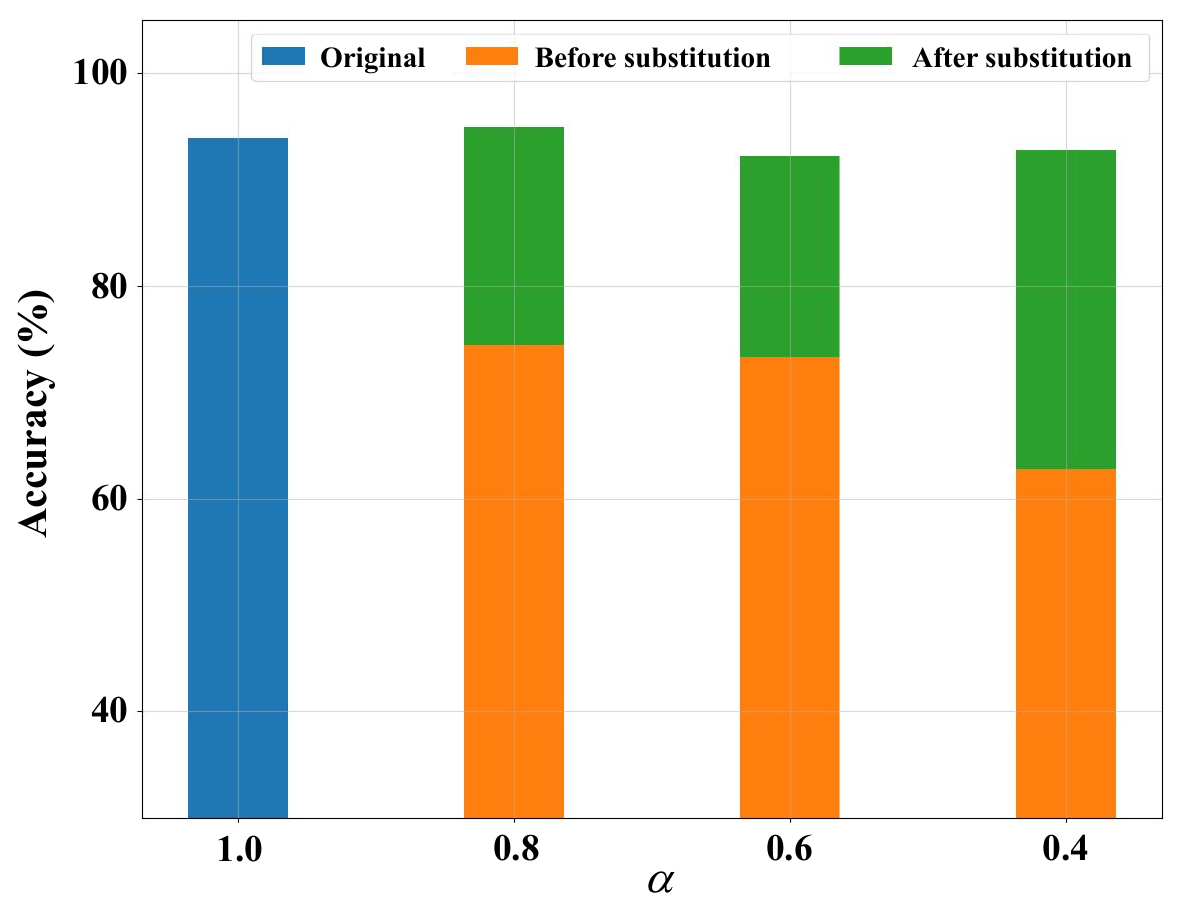}
    \caption{
    Recognition performance before and after data substitution. The symbol $\alpha$ denotes the proportion of real defect images utilized for training. The blue bar shows performance with all real images used, the orange bar indicates performance with only a subset of real images, and the green bar reflects performance improvements when $(1 - \alpha)$ percent of the real images are substituted with generated ones.
    }
    \label{fig_classification_NEU}
\end{figure}

\textbf{Data substitution}. To further prove that our generated data and collected samples have high distribution overlap, we compare the recognition performance of the models trained on the datasets before and after the data substitution. 
The NEU dataset is divided into three portions: the training subset, the validation subset, and the testing sets, with the division being in the ratio of 8:1:1.
Subsequently, we proceed to train the recognition model using only a fraction of the training subset. Let $\alpha$ represent the fraction of the original real images that are used for training, with values set to 0.8, 0.6, and 0.4, respectively.
We use SqueezeNet~\cite{iandola2016squeezenet} as the recognition model, and the evaluation results are shown in Fig.~\ref{fig_classification_NEU}.
It is not difficult to find that, a reduction in $\alpha$ corresponds to fewer training samples, which causes a corresponding decrease in the accuracy of the recognition model. 
This trend highlights the importance of the amount of training data for the performance of the recognition model.
Then we supplement the training data with generated samples to match the original quantity of training subset, the performance of the recognition model can achieve the comparable accuracy to that obtained when using the complete original training subset, proving the effectiveness of the defect samples generated by the proposed method.

\begin{table}[!t]
	\setlength{\tabcolsep}{0.3cm}
    \renewcommand\arraystretch{1.2}
	\begin{center}
	\caption{
	Quantitative experimental results for \\ defect recognition on NEU.
	}
	\label{classification_NEU}
    \footnotesize
	\begin{tabular}{ccc}
	\hline\hline    
    Networks&Expansion Methods&Accuracy(\%)\\
    \hline
\multirow{8}*{AlexNet}&None&81.67\\
&{DCGAN}&92.22\\
&{StyleGAN3}&94.44\\
&{DDIM}&96.67\\
&{Textual Inversion}&{91.67}\\
&{DreamBooth}&{92.22}\\
&{Yang~\textit{et al.}}&{96.11}\\
&StableSDG&\textbf{98.33}\\
\hline
\multirow{8}*{VGG}&None&86.67\\
&{DCGAN}&89.44\\
&{StyleGAN3}&93.89\\
&{DDIM}&96.67\\
&{Textual Inversion}&{90.00}\\
&{DreamBooth}&{95.56}\\
&{Yang~\textit{et al.}}&{97.78}\\
&StableSDG&\textbf{98.33}\\
\hline
\multirow{8}*{ResNet}&None&77.78\\
&{DCGAN}&87.22\\
&{StyleGAN3}&91.67\\
&{DDIM}&95.00\\
&{Textual Inversion}&{92.78}\\
&{DreamBooth}&{93.89}\\
&{Yang~\textit{et al.}}&{95.56}\\
&StableSDG&\textbf{97.78}\\
\hline
\multirow{8}*{SqueezeNet}&None&62.78\\
&{DCGAN}&79.44\\
&{StyleGAN3}&93.89\\
&{DDIM}&96.11\\
&{Textual Inversion}&{90.56}\\
&{DreamBooth}&{91.11}\\
&{Yang~\textit{et al.}}&{96.11}\\
&StableSDG&\textbf{97.78}\\
\hline
\multirow{8}*{DenseNet}&None&73.33\\
&{DCGAN}&80.00\\
&{StyleGAN3}&91.67\\
&{DDIM}&92.78\\
&{Textual Inversion}&{85.56}\\
&{DreamBooth}&{92.78}\\
&{Yang~\textit{et al.}}&{92.78}\\
&StableSDG&\textbf{94.44}\\
\hline\hline    
\end{tabular}
\end{center}
\end{table}

\textbf{Dataset expansion}. Considering that replacing the real samples in the training subset with generated ones brings the result nearly equal to the original accuracy, we introduce an additional 1,000 generated samples to the training subset within the $\alpha=0.4$ configuration (where each class has 96 real images) to determine if there can be further enhancements to the recognition performance of model.
This experiment is conducted with multiple off-the-shelf network architectures, i.e., AlexNet~\cite{krizhevsky2012imagenet}, VGG~\cite{simonyan2014very}, ResNet~\cite{he2016deep}, SqueezeNet~\cite{iandola2016squeezenet} and DenseNet~\cite{huang2017densely}.
According to Table~\ref{classification_NEU}, there are consistent performance improvement when the generated images are introduced by all methods, showing that expanding the defect dataset with generative model is significantly helpful for higher defect recognition accuracy.
When compared with other data expansion methods, StableSDG exhibits a superior ability to enhance the performance of steel surface defect recognition. The accuracy of the five recognition models is, on average, improved by approximately 20\%, demonstrating that our method can expand the dataset more effectively.

\begin{table}[!t]
	\setlength{\tabcolsep}{0.04cm}
    \renewcommand\arraystretch{1.2}
	\begin{center}
	\caption{
	Quantitative comparison among various image generative models trained on CCBSD.
	}
	\label{quality_CCBSD}
    \footnotesize
	\begin{tabular}{c|cccc}
	\hline\hline
	\multirow{2}*{}&\multicolumn{4}{c}{FID Scores $\downarrow$}\\   
    &Inclusion&Indentation&Oxidation&Slag Groove\\
    \hline
{DCGAN \textcolor{gray}{\scriptsize{(CCC~18~\cite{li2018unsupervised})}}}&328.58&270.33&277.52&242.12\\
{StyleGAN3 \textcolor{gray}{\scriptsize{(NIPS~21~\cite{karras2021alias})}}}&167.06&157.19&99.08&175.54\\
{DDIM \textcolor{gray}{\scriptsize{(ICLR~21~\cite{song2020denoising})}}}&166.76&176.14&81.60&113.24\\
{{Textual Inversion~\textcolor{gray}{\scriptsize{(ICLR~23~\cite{gal2022image})}}}}&{257.31}&{146.15}&{282.77}&{202.93}\\
{{DreamBooth~\textcolor{gray}{\scriptsize{(CVPR~23~\cite{ruiz2023dreambooth})}}}}&{206.12}&{137.22}&{130.85}&{235.51}\\
{{Yang~\textit{et al.}~\textcolor{gray}{\scriptsize{(TII~24~\cite{yang2024novel})}}}}&{165.78}&{161.19}&{102.54}&{127.23}\\
{StableSDG (Ours)}&\textbf{112.68}&\textbf{72.18}&\textbf{71.79}&\textbf{97.13}\\
\hline\hline
\end{tabular}
\end{center}
\end{table}

\begin{figure}[!t] 
    \centering
    \includegraphics[scale=0.48]{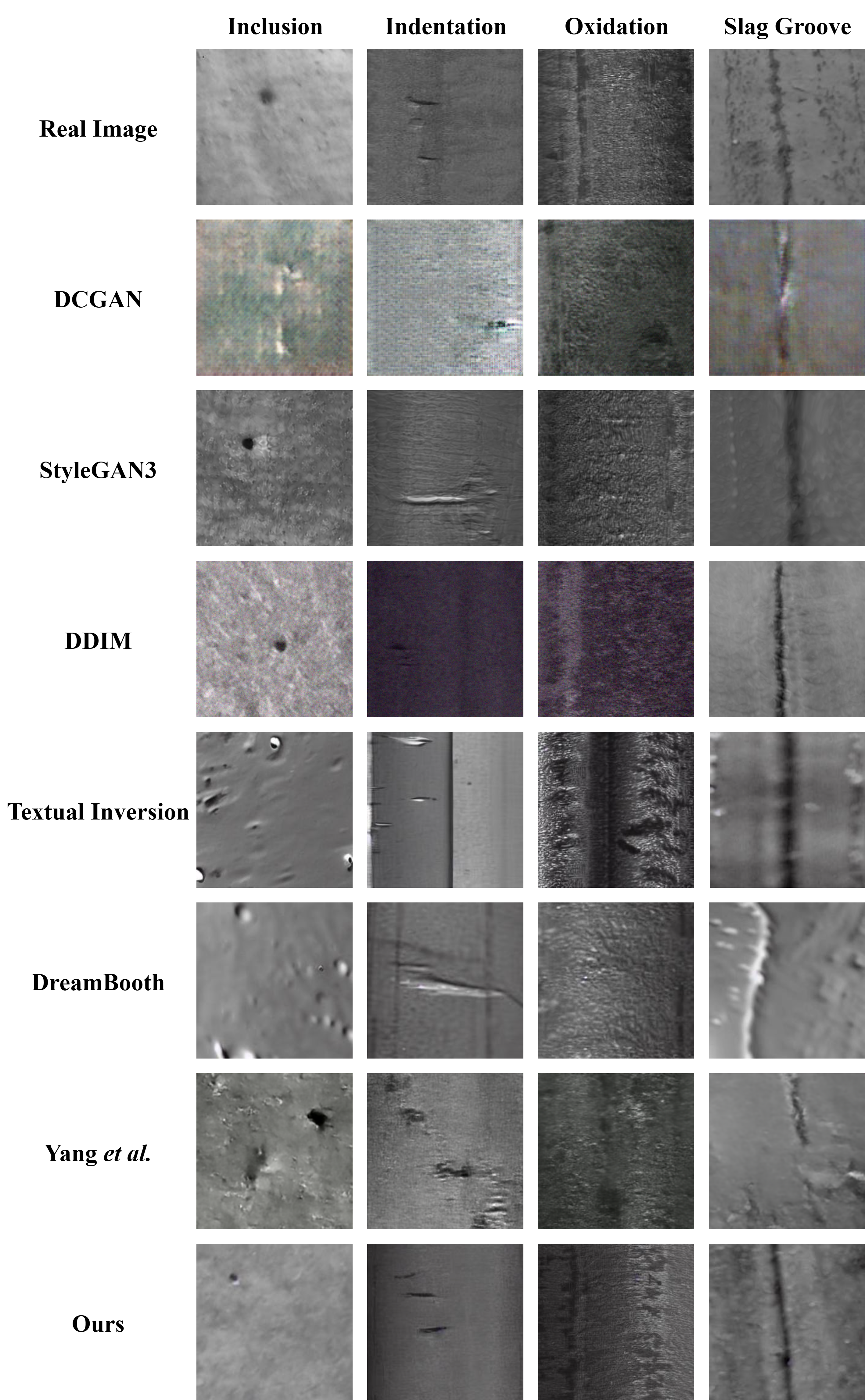}
    \caption{Qualitative comparison of StableSDG with other models on CCBSD.}
    \label{visualization_CCBSD}
\end{figure}

\begin{table}[!t]
	\setlength{\tabcolsep}{0.3cm}
    \renewcommand\arraystretch{1.2}
	\begin{center}
	\caption{
	Quantitative experimental results for \\ defect recognition on CCBSD.
	}
	\label{classification_CCBSD}
    \footnotesize
	\begin{tabular}{ccc}
	\hline\hline    
    Networks&Expansion Methods&Accuracy(\%)\\
    \hline
\multirow{8}*{AlexNet}&None&82.50\\
&{DCGAN}&92.50\\
&{StyleGAN3}&95.00\\
&{DDIM}&94.17\\
&{Textual Inversion}&{93.33}\\
&{DreamBooth}&{93.33}\\
&{Yang~\textit{et al.}}&{95.83}\\
&StableSDG&\textbf{98.33}\\
\hline
\multirow{8}*{VGG}&None&80.00\\
&{DCGAN}&93.33\\
&{StyleGAN3}&95.83\\
&{DDIM}&96.67\\
&{Textual Inversion}&{94.17}\\
&{DreamBooth}&{95.00}\\
&{Yang~\textit{et al.}}&{95.83}\\
&StableSDG&\textbf{98.33}\\
\hline
\multirow{8}*{ResNet}&None&88.33\\
&{DCGAN}&95.00\\
&{StyleGAN3}&96.67\\
&{DDIM}&95.83\\
&{Textual Inversion}&{95.83}\\
&{DreamBooth}&{96.67}\\
&{Yang~\textit{et al.}}&{96.67}\\
&StableSDG&\textbf{99.17}\\
\hline
\multirow{8}*{SqueezeNet}&None&85.00\\
&{DCGAN}&88.33\\
&{StyleGAN3}&89.17\\
&{DDIM}&89.17\\
&{Textual Inversion}&{87.50}\\
&{DreamBooth}&{90.00}\\
&{Yang~\textit{et al.}}&{90.00}\\
&StableSDG&\textbf{91.67}\\
\hline
\multirow{8}*{DenseNet}&None&90.00\\
&{DCGAN}&93.33\\
&{StyleGAN3}&97.50\\
&{DDIM}&98.33\\
&{Textual Inversion}&{93.33}\\
&{DreamBooth}&{95.00}\\
&{Yang~\textit{et al.}}&{96.67}\\
&StableSDG&\textbf{99.17}\\
\hline\hline    
\end{tabular}
\end{center}
\end{table}

\subsection{Performance on CCBSD}
\label{performance_CCBSD}

\textbf{Generation quality}. To further evaluate StableSDG, we also conduct experiments on CCBSD.
The dataset contains four categories of defects with 200 samples per category, 140 for training, 30 for validation, and 30 for testing. 
The quantitative comparison of defect image generation is shown in Table~\ref{quality_CCBSD}. It shows that StableSDG can achieve lower FID scores. For qualitative comparison, Fig.~\ref{visualization_CCBSD} represents the images generated by StableSDG and comparisons with other generative methods.
We can see that the samples generated by DCGAN show artificial textures in the background. Textual Inversion and DreamBooth have enhanced the quality of their generated images, though some atypical patterns persist. Meanwhile, the samples from StyleGAN3, DDIM, and Yang \textit{et al.} appear visually similar to real images but suffer from some blurriness. In contrast, the samples generated by StableSDG display well-defined defect characteristics. This indicates the superiority of our method in data expansion of continuous casting billet surface defect samples.

\textbf{Dataset expansion}. For each defect category, 1,000 samples are generated by different methods respectively. Then we adopt the aforementioned recognition models to perform the continuous casting billet surface defect recognition on the dataset before and after expansion.
The experimental results are shown in Table~\ref{classification_CCBSD}.
It can be found that, the accuracy of recognition models is higher after adding generated samples to the original training set.
Compared to the other expansion methods, StableSDG has greater advantages in improving the performance of continuous casting billet surface defect recognition.
The five recognition models experience an average accuracy improvement of approximately 12\%, which confirms the effectiveness of our method in expanding the dataset. This enhancement also enables the recognition models to be further utilized for recognizing surface defects in continuous casting billets within industrial manufacturing settings.

\section{Conclusion}
\label{conclusion}
The scarcity of data samples presents a significant challenge for deploying deep learning techniques in the recognition of steel surface defects. To address this problem, we introduce StableSDG, which blends text-to-image prior for defect image generation.
During the process of generator adaptation, StableSDG adapts and modifies within both the token embedding space and the network parameters space. 
When generating data, it generates samples from image-oriented initialization, instead of starting from pure Gaussian noises.
The experimental results on NEU and CCBSD verify that the proposed method can generate defect images with high fidelity, which can greatly improve the performance of recognition models.
However, the proposed method is limited to text prompts, resulting in image generation that is stochastic and lacks direction.
In the future, we plan to explore using other modalities as conditions, e.g., spatial conditions, to generate images adhering to the spatial conditioning input. In doing so, we can generate defect sample images that include bounding box information, which can be leveraged to improve the performance of neural networks in defect detection tasks.

\ifCLASSOPTIONcaptionsoff
  \newpage
\fi

\bibliographystyle{IEEEtran}
\bibliography{reference}

\begin{thebibliography}{10}
\providecommand{\url}[1]{#1}
\csname url@samestyle\endcsname
\providecommand{\newblock}{\relax}
\providecommand{\bibinfo}[2]{#2}
\providecommand{\BIBentrySTDinterwordspacing}{\spaceskip=0pt\relax}
\providecommand{\BIBentryALTinterwordstretchfactor}{4}
\providecommand{\BIBentryALTinterwordspacing}{\spaceskip=\fontdimen2\font plus
\BIBentryALTinterwordstretchfactor\fontdimen3\font minus \fontdimen4\font\relax}
\providecommand{\BIBforeignlanguage}[2]{{%
\expandafter\ifx\csname l@#1\endcsname\relax
\typeout{** WARNING: IEEEtran.bst: No hyphenation pattern has been}%
\typeout{** loaded for the language `#1'. Using the pattern for}%
\typeout{** the default language instead.}%
\else
\language=\csname l@#1\endcsname
\fi
#2}}
\providecommand{\BIBdecl}{\relax}
\BIBdecl

\bibitem{ghorai2012automatic}
S.~Ghorai, A.~Mukherjee, M.~Gangadaran, and P.~K. Dutta, ``Automatic defect detection on hot-rolled flat steel products,'' \emph{IEEE Transactions on Instrumentation and Measurement}, vol.~62, no.~3, pp. 612--621, 2012.

\bibitem{hassan2019underground}
S.~I. Hassan, L.~M. Dang, I.~Mehmood, S.~Im, C.~Choi, J.~Kang, Y.-S. Park, and H.~Moon, ``Underground sewer pipe condition assessment based on convolutional neural networks,'' \emph{Automation in Construction}, vol. 106, p. 102849, 2019.

\bibitem{ai2013surface}
Y.-h. Ai and K.~Xu, ``Surface detection of continuous casting slabs based on curvelet transform and kernel locality preserving projections,'' \emph{Journal of Iron and Steel Research International}, vol.~20, no.~5, pp. 80--86, 2013.

\bibitem{choi2014algorithm}
D.-C. Choi, Y.-J. Jeon, S.~J. Lee, J.~P. Yun, and S.~W. Kim, ``Algorithm for detecting seam cracks in steel plates using a gabor filter combination method,'' \emph{Applied optics}, vol.~53, no.~22, pp. 4865--4872, 2014.

\bibitem{dongyan2016defect}
C.~Dongyan, X.~Kewen, N.~Aslam, and H.~Jingzhong, ``Defect classification recognition on strip steel surface using second-order cone programming-relevance vector machine algorithm,'' \emph{Journal of Computational and Theoretical Nanoscience}, vol.~13, no.~9, pp. 6141--6148, 2016.

\bibitem{cheon2019convolutional}
S.~Cheon, H.~Lee, C.~O. Kim, and S.~H. Lee, ``Convolutional neural network for wafer surface defect classification and the detection of unknown defect class,'' \emph{IEEE Transactions on Semiconductor Manufacturing}, vol.~32, no.~2, pp. 163--170, 2019.

\bibitem{konovalenko2020steel}
I.~Konovalenko, P.~Maruschak, J.~Brezinov{\'a}, J.~Vi{\v{n}}{\'a}{\v{s}}, and J.~Brezina, ``Steel surface defect classification using deep residual neural network,'' \emph{Metals}, vol.~10, no.~6, p. 846, 2020.

\bibitem{wang2021new}
Y.~Wang, L.~Gao, Y.~Gao, and X.~Li, ``A new graph-based semi-supervised method for surface defect classification,'' \emph{Robotics and Computer-Integrated Manufacturing}, vol.~68, p. 102083, 2021.

\bibitem{yun2020automated}
J.~P. Yun, W.~C. Shin, G.~Koo, M.~S. Kim, C.~Lee, and S.~J. Lee, ``Automated defect inspection system for metal surfaces based on deep learning and data augmentation,'' \emph{Journal of Manufacturing Systems}, vol.~55, pp. 317--324, 2020.

\bibitem{niu2020defect}
S.~Niu, B.~Li, X.~Wang, and H.~Lin, ``Defect image sample generation with gan for improving defect recognition,'' \emph{IEEE Transactions on Automation Science and Engineering}, vol.~17, no.~3, pp. 1611--1622, 2020.

\bibitem{zhao2023defect}
C.~Zhao, W.~Xue, W.~Fu, Z.~Li, and X.~Fang, ``Defect sample image generation method based on gans in diamond tool defect detection,'' \emph{IEEE Transactions on Instrumentation and Measurement}, 2023.

\bibitem{zhang2022diversifying}
Y.~Zhang, Y.~Wang, Z.~Jiang, F.~Liao, L.~Zheng, D.~Tan, J.~Chen, and J.~Lu, ``Diversifying tire-defect image generation based on generative adversarial network,'' \emph{IEEE Transactions on Instrumentation and Measurement}, vol.~71, pp. 1--12, 2022.

\bibitem{zhang2021defect}
G.~Zhang, K.~Cui, T.-Y. Hung, and S.~Lu, ``Defect-gan: High-fidelity defect synthesis for automated defect inspection,'' in \emph{Proceedings of the IEEE/CVF Winter Conference on Applications of Computer Vision}, 2021, pp. 2524--2534.

\bibitem{duan2023few}
Y.~Duan, Y.~Hong, L.~Niu, and L.~Zhang, ``Few-shot defect image generation via defect-aware feature manipulation,'' in \emph{Proceedings of the AAAI Conference on Artificial Intelligence}, vol.~37, no.~1, 2023, pp. 571--578.

\bibitem{li2023dls}
W.~Li, C.~Gu, J.~Chen, C.~Ma, X.~Zhang, B.~Chen, and S.~Wan, ``Dls-gan: generative adversarial nets for defect location sensitive data augmentation,'' \emph{IEEE Transactions on Automation Science and Engineering}, 2023.

\bibitem{ding2021cogview}
M.~Ding, Z.~Yang, W.~Hong, W.~Zheng, C.~Zhou, D.~Yin, J.~Lin, X.~Zou, Z.~Shao, H.~Yang \emph{et~al.}, ``Cogview: Mastering text-to-image generation via transformers,'' \emph{Advances in Neural Information Processing Systems}, vol.~34, pp. 19\,822--19\,835, 2021.

\bibitem{ramesh2021zero}
A.~Ramesh, M.~Pavlov, G.~Goh, S.~Gray, C.~Voss, A.~Radford, M.~Chen, and I.~Sutskever, ``Zero-shot text-to-image generation,'' in \emph{International Conference on Machine Learning}.\hskip 1em plus 0.5em minus 0.4em\relax PMLR, 2021, pp. 8821--8831.

\bibitem{nichol2021glide}
A.~Nichol, P.~Dhariwal, A.~Ramesh, P.~Shyam, P.~Mishkin, B.~McGrew, I.~Sutskever, and M.~Chen, ``Glide: Towards photorealistic image generation and editing with text-guided diffusion models,'' \emph{arXiv preprint arXiv:2112.10741}, 2021.

\bibitem{balaji2022ediffi}
Y.~Balaji, S.~Nah, X.~Huang, A.~Vahdat, J.~Song, K.~Kreis, M.~Aittala, T.~Aila, S.~Laine, B.~Catanzaro \emph{et~al.}, ``ediffi: Text-to-image diffusion models with an ensemble of expert denoisers,'' \emph{arXiv preprint arXiv:2211.01324}, 2022.

\bibitem{ramesh2022hierarchical}
A.~Ramesh, P.~Dhariwal, A.~Nichol, C.~Chu, and M.~Chen, ``Hierarchical text-conditional image generation with clip latents,'' \emph{arXiv preprint arXiv:2204.06125}, vol.~1, no.~2, p.~3, 2022.

\bibitem{saharia2022photorealistic}
C.~Saharia, W.~Chan, S.~Saxena, L.~Li, J.~Whang, E.~L. Denton, K.~Ghasemipour, R.~Gontijo~Lopes, B.~Karagol~Ayan, T.~Salimans \emph{et~al.}, ``Photorealistic text-to-image diffusion models with deep language understanding,'' \emph{Advances in Neural Information Processing Systems}, vol.~35, pp. 36\,479--36\,494, 2022.

\bibitem{rombach2022high}
R.~Rombach, A.~Blattmann, D.~Lorenz, P.~Esser, and B.~Ommer, ``High-resolution image synthesis with latent diffusion models,'' in \emph{Proceedings of the IEEE/CVF conference on computer vision and pattern recognition}, 2022, pp. 10\,684--10\,695.

\bibitem{wang2023exploiting}
J.~Wang, Z.~Yue, S.~Zhou, K.~C. Chan, and C.~C. Loy, ``Exploiting diffusion prior for real-world image super-resolution,'' \emph{arXiv preprint arXiv:2305.07015}, 2023.

\bibitem{xia2023diffir}
B.~Xia, Y.~Zhang, S.~Wang, Y.~Wang, X.~Wu, Y.~Tian, W.~Yang, and L.~Van~Gool, ``Diffir: Efficient diffusion model for image restoration,'' \emph{arXiv preprint arXiv:2303.09472}, 2023.

\bibitem{kawar2023imagic}
B.~Kawar, S.~Zada, O.~Lang, O.~Tov, H.~Chang, T.~Dekel, I.~Mosseri, and M.~Irani, ``Imagic: Text-based real image editing with diffusion models,'' in \emph{Proceedings of the IEEE/CVF Conference on Computer Vision and Pattern Recognition}, 2023, pp. 6007--6017.

\bibitem{gal2022image}
R.~Gal, Y.~Alaluf, Y.~Atzmon, O.~Patashnik, A.~H. Bermano, G.~Chechik, and D.~Cohen-Or, ``An image is worth one word: Personalizing text-to-image generation using textual inversion,'' \emph{arXiv preprint arXiv:2208.01618}, 2022.

\bibitem{ruiz2023dreambooth}
N.~Ruiz, Y.~Li, V.~Jampani, Y.~Pritch, M.~Rubinstein, and K.~Aberman, ``Dreambooth: Fine tuning text-to-image diffusion models for subject-driven generation,'' in \emph{Proceedings of the IEEE/CVF Conference on Computer Vision and Pattern Recognition}, 2023, pp. 22\,500--22\,510.

\bibitem{han2023svdiff}
L.~Han, Y.~Li, H.~Zhang, P.~Milanfar, D.~Metaxas, and F.~Yang, ``Svdiff: Compact parameter space for diffusion fine-tuning,'' \emph{arXiv preprint arXiv:2303.11305}, 2023.

\bibitem{chen2023subject}
W.~Chen, H.~Hu, Y.~Li, N.~Rui, X.~Jia, M.-W. Chang, and W.~W. Cohen, ``Subject-driven text-to-image generation via apprenticeship learning,'' \emph{arXiv preprint arXiv:2304.00186}, 2023.

\bibitem{chen2023disenbooth}
H.~Chen, Y.~Zhang, X.~Wang, X.~Duan, Y.~Zhou, and W.~Zhu, ``Disenbooth: Disentangled parameter-efficient tuning for subject-driven text-to-image generation,'' \emph{arXiv preprint arXiv:2305.03374}, 2023.

\bibitem{hu2021lora}
E.~J. Hu, Y.~Shen, P.~Wallis, Z.~Allen-Zhu, Y.~Li, S.~Wang, L.~Wang, and W.~Chen, ``Lora: Low-rank adaptation of large language models,'' \emph{arXiv preprint arXiv:2106.09685}, 2021.

\bibitem{song2022coarse}
Y.~Song, Z.~Liu, S.~Ling, R.~Tang, G.~Duan, and J.~Tan, ``Coarse-to-fine few-shot defect recognition with dynamic weighting and joint metric,'' \emph{IEEE Transactions on Instrumentation and Measurement}, vol.~71, pp. 1--10, 2022.

\bibitem{wang2022graph}
Y.~Wang, L.~Gao, Y.~Gao, and X.~Li, ``A graph guided convolutional neural network for surface defect recognition,'' \emph{IEEE Transactions on Automation Science and Engineering}, vol.~19, no.~3, pp. 1392--1404, 2022.

\bibitem{wang2023few}
T.~Wang, Z.~Li, Y.~Xu, J.~Chen, A.~Genovese, V.~Piuri, and F.~Scotti, ``Few-shot steel surface defect recognition via self-supervised teacher-student model with min-max instances similarity,'' \emph{IEEE Transactions on Instrumentation and Measurement}, 2023.

\bibitem{dong2021defect}
X.~Dong, C.~J. Taylor, and T.~F. Cootes, ``Defect classification and detection using a multitask deep one-class cnn,'' \emph{IEEE Transactions on Automation Science and Engineering}, vol.~19, no.~3, pp. 1719--1730, 2021.

\bibitem{mo2020weighted}
D.~Mo, W.~K. Wong, Z.~Lai, and J.~Zhou, ``Weighted double-low-rank decomposition with application to fabric defect detection,'' \emph{IEEE Transactions on Automation Science and Engineering}, vol.~18, no.~3, pp. 1170--1190, 2020.

\bibitem{zhang2023duak}
Y.~Zhang, H.~Wang, W.~Shen, and G.~Peng, ``Duak: Reinforcement learning-based knowledge graph reasoning for steel surface defect detection,'' \emph{IEEE Transactions on Automation Science and Engineering}, 2023.

\bibitem{huang2009template}
Q.~Huang, Y.~Wu, J.~Baruch, P.~Jiang, and Y.~Peng, ``A template model for defect simulation for evaluating nondestructive testing in x-radiography,'' \emph{IEEE Transactions on Systems, Man, and Cybernetics-Part A: Systems and Humans}, vol.~39, no.~2, pp. 466--475, 2009.

\bibitem{mery2005simulation}
D.~Mery, D.~Hahn, and N.~Hitschfeld, ``Simulation of defects in aluminium castings using cad models of flaws and real x-ray images,'' \emph{Insight-Non-Destructive Testing and Condition Monitoring}, vol.~47, no.~10, pp. 618--624, 2005.

\bibitem{mery2002automated}
D.~Mery and D.~Filbert, ``Automated flaw detection in aluminum castings based on the tracking of potential defects in a radioscopic image sequence,'' \emph{IEEE Transactions on Robotics and Automation}, vol.~18, no.~6, pp. 890--901, 2002.

\bibitem{kingma2013auto}
D.~P. Kingma and M.~Welling, ``Auto-encoding variational bayes,'' \emph{arXiv preprint arXiv:1312.6114}, 2013.

\bibitem{goodfellow2014generative}
I.~Goodfellow, J.~Pouget-Abadie, M.~Mirza, B.~Xu, D.~Warde-Farley, S.~Ozair, A.~Courville, and Y.~Bengio, ``Generative adversarial nets,'' \emph{Advances in neural information processing systems}, vol.~27, 2014.

\bibitem{li2018unsupervised}
J.~Li, J.~Jia, and D.~Xu, ``Unsupervised representation learning of image-based plant disease with deep convolutional generative adversarial networks,'' in \emph{2018 37th Chinese control conference (CCC)}.\hskip 1em plus 0.5em minus 0.4em\relax IEEE, 2018, pp. 9159--9163.

\bibitem{karras2021alias}
T.~Karras, M.~Aittala, S.~Laine, E.~H{\"a}rk{\"o}nen, J.~Hellsten, J.~Lehtinen, and T.~Aila, ``Alias-free generative adversarial networks,'' \emph{Advances in Neural Information Processing Systems}, vol.~34, pp. 852--863, 2021.

\bibitem{sauer2022stylegan}
A.~Sauer, K.~Schwarz, and A.~Geiger, ``Stylegan-xl: Scaling stylegan to large diverse datasets,'' in \emph{ACM SIGGRAPH 2022 conference proceedings}, 2022, pp. 1--10.

\bibitem{yang2024novel}
X.~Yang, T.~Ye, X.~Yuan, W.~Zhu, X.~Mei, and F.~Zhou, ``A novel data augmentation method based on denoising diffusion probabilistic model for fault diagnosis under imbalanced data,'' \emph{IEEE Transactions on Industrial Informatics}, 2024.

\bibitem{ho2020denoising}
J.~Ho, A.~Jain, and P.~Abbeel, ``Denoising diffusion probabilistic models,'' \emph{Advances in neural information processing systems}, vol.~33, pp. 6840--6851, 2020.

\bibitem{song2020denoising}
J.~Song, C.~Meng, and S.~Ermon, ``Denoising diffusion implicit models,'' \emph{arXiv preprint arXiv:2010.02502}, 2020.

\bibitem{zhang2024compositional}
X.~Zhang, X.-Y. Wei, J.~Wu, T.~Zhang, Z.~Zhang, Z.~Lei, and Q.~Li, ``Compositional inversion for stable diffusion models,'' in \emph{Proceedings of the AAAI Conference on Artificial Intelligence}, vol.~38, no.~7, 2024, pp. 7350--7358.

\bibitem{cai2024decoupled}
Y.~Cai, Y.~Wei, Z.~Ji, J.~Bai, H.~Han, and W.~Zuo, ``Decoupled textual embeddings for customized image generation,'' in \emph{Proceedings of the AAAI Conference on Artificial Intelligence}, vol.~38, no.~2, 2024, pp. 909--917.

\bibitem{kumari2023multi}
N.~Kumari, B.~Zhang, R.~Zhang, E.~Shechtman, and J.-Y. Zhu, ``Multi-concept customization of text-to-image diffusion,'' in \emph{Proceedings of the IEEE/CVF Conference on Computer Vision and Pattern Recognition}, 2023, pp. 1931--1941.

\bibitem{chen2024subject}
W.~Chen, H.~Hu, Y.~Li, N.~Ruiz, X.~Jia, M.-W. Chang, and W.~W. Cohen, ``Subject-driven text-to-image generation via apprenticeship learning,'' \emph{Advances in Neural Information Processing Systems}, vol.~36, 2024.

\bibitem{ronneberger2015u}
O.~Ronneberger, P.~Fischer, and T.~Brox, ``U-net: Convolutional networks for biomedical image segmentation,'' in \emph{Medical image computing and computer-assisted intervention--MICCAI 2015: 18th international conference, Munich, Germany, October 5-9, 2015, proceedings, part III 18}.\hskip 1em plus 0.5em minus 0.4em\relax Springer, 2015, pp. 234--241.

\bibitem{radford2021learning}
A.~Radford, J.~W. Kim, C.~Hallacy, A.~Ramesh, G.~Goh, S.~Agarwal, G.~Sastry, A.~Askell, P.~Mishkin, J.~Clark \emph{et~al.}, ``Learning transferable visual models from natural language supervision,'' in \emph{International conference on machine learning}.\hskip 1em plus 0.5em minus 0.4em\relax PMLR, 2021, pp. 8748--8763.

\bibitem{vaswani2017attention}
A.~Vaswani, N.~Shazeer, N.~Parmar, J.~Uszkoreit, L.~Jones, A.~N. Gomez, {\L}.~Kaiser, and I.~Polosukhin, ``Attention is all you need,'' \emph{Advances in neural information processing systems}, vol.~30, 2017.

\bibitem{ho2022classifier}
J.~Ho and T.~Salimans, ``Classifier-free diffusion guidance,'' \emph{arXiv preprint arXiv:2207.12598}, 2022.

\bibitem{heusel2017gans}
M.~Heusel, H.~Ramsauer, T.~Unterthiner, B.~Nessler, and S.~Hochreiter, ``Gans trained by a two time-scale update rule converge to a local nash equilibrium,'' \emph{Advances in neural information processing systems}, vol.~30, 2017.

\bibitem{krizhevsky2012imagenet}
A.~Krizhevsky, I.~Sutskever, and G.~E. Hinton, ``Imagenet classification with deep convolutional neural networks,'' \emph{Advances in neural information processing systems}, vol.~25, 2012.

\bibitem{simonyan2014very}
K.~Simonyan and A.~Zisserman, ``Very deep convolutional networks for large-scale image recognition,'' \emph{arXiv preprint arXiv:1409.1556}, 2014.

\bibitem{he2016deep}
K.~He, X.~Zhang, S.~Ren, and J.~Sun, ``Deep residual learning for image recognition,'' in \emph{Proceedings of the IEEE conference on computer vision and pattern recognition}, 2016, pp. 770--778.

\bibitem{iandola2016squeezenet}
F.~N. Iandola, S.~Han, M.~W. Moskewicz, K.~Ashraf, W.~J. Dally, and K.~Keutzer, ``Squeezenet: Alexnet-level accuracy with 50x fewer parameters and< 0.5 mb model size,'' \emph{arXiv preprint arXiv:1602.07360}, 2016.

\bibitem{huang2017densely}
G.~Huang, Z.~Liu, L.~Van Der~Maaten, and K.~Q. Weinberger, ``Densely connected convolutional networks,'' in \emph{Proceedings of the IEEE conference on computer vision and pattern recognition}, 2017, pp. 4700--4708.

\bibitem{song2013noise}
K.~Song and Y.~Yan, ``A noise robust method based on completed local binary patterns for hot-rolled steel strip surface defects,'' \emph{Applied Surface Science}, vol. 285, pp. 858--864, 2013.

\bibitem{stable-diffusion-v1-5}
R.~Rombach, A.~Blattmann, D.~Lorenz, P.~Esser, and B.~Ommer, ``Stable-diffusion-v1-5,'' \url{https://huggingface.co/runwayml/stable-diffusion-v1-5}.

\bibitem{kingma2014adam}
D.~P. Kingma and J.~Ba, ``Adam: A method for stochastic optimization,'' \emph{arXiv preprint arXiv:1412.6980}, 2014.

\end{thebibliography}

\end{document}